\documentclass[sigconf, conference]{acmart}

\usepackage{multirow}
\usepackage{caption}
\usepackage{booktabs}
\usepackage{colortbl}
\usepackage{tabularx}
\usepackage{xcolor}
\usepackage{float}
\usepackage[normalem]{ulem}
\usepackage[many]{tcolorbox}    	
\usepackage{subcaption}
\usepackage{listings}
\definecolor{quartz}{HTML}{D2D2D9}
\definecolor{lavender}{HTML}{EAEAF2}
\definecolor{mint}{HTML}{D0F0C0}

\definecolor{codegreen}{rgb}{0,0.6,0}
\definecolor{codegray}{rgb}{0.5,0.5,0.5}
\definecolor{codepurple}{rgb}{0.58,0,0.82}

\colorlet{tableheadcolor}{lavender} 
\colorlet{tablerowcolor}{lavender} 
\colorlet{mintrowcolor}{mint} 
\newcommand{\headcol}{\rowcolor{tableheadcolor}}
\newcommand{\rowcol}{\rowcolor{tablerowcolor}}
\newcommand{\mintrow}{\rowcolor{mintrowcolor}}

\newcommand{\midsepremove}{\aboverulesep = 0mm \belowrulesep = 0mm}

\newcommand{\midsepdefault}{\aboverulesep = 0.605mm \belowrulesep = 0.984mm}

\tcbset{
    sharp corners,
    before skip = 0.25cm,    
    after skip = 0.25cm      
}                           

\newtcolorbox{boxC}{
    colback = tablerowcolor, 
    colframe = quartz, 
    boxrule = 0pt,  
    toprule = 3pt
}

\lstdefinelanguage{json}{
    string=[s]{"}{"},
    stringstyle=\color{blue},
    comment=[l]{:},
    commentstyle=\color{black},
    numberstyle=\tiny\color{codegray}
}

\lstdefinestyle{jsonstyle}{
    backgroundcolor=\color{lavender},   
    commentstyle=\color{codegreen},
    keywordstyle=\color{magenta},
    numberstyle=\tiny\color{codegray},
    stringstyle=\color{codepurple},
    basicstyle=\ttfamily\small,
    breakatwhitespace=false,
    breaklines=true,
    captionpos=b,
    keepspaces=true,
    numbersep=5pt,
    showspaces=false,
    showstringspaces=false,
    showtabs=false,
    tabsize=2
}

\AtBeginDocument{%
  \providecommand\BibTeX{{%
    \normalfont B\kern-0.5em{\scshape i\kern-0.25em b}\kern-0.8em\TeX}}}

\setcopyright{acmlicensed}
\copyrightyear{2018}
\acmYear{2018}
\acmDOI{XXXXXXX.XXXXXXX}

\acmConference[ISSTA 2024]{ACM SIGSOFT International Symposium on Software Testing and Analysis}{16-20 September, 2024}{Vienna, Austria}
\acmISBN{978-1-4503-XXXX-X/18/06}

\begin{document}

\title{Evaluating Deep Neural Networks in Deployment: A Comparative Study (Replicability Study)}

\author{Eduard Pinconschi}
\email{epincons@andrew.cmu.edu}
\orcid{0000-0002-0729-4684}
\affiliation{%
  \institution{Carnegie Mellon University}
  \city{Pittsburgh}
  \state{Pennsylvania}
  \country{USA}
}

\author{Divya Gopinath}
\email{divya.gopinath@nasa.gov}
\orcid{0000-0002-1242-7701}
\affiliation{%
  \institution{KBR Inc., NASA Ames}
  \city{Mountain View}
  \state{California}
  \country{USA}
}

\author{Rui Abreu}
\email{rui@computer.org}
\orcid{0000-0003-3734-3157}
\affiliation{%
  \institution{INESC-ID, University of Porto}
  \city{Porto}
  \country{Portugal}
}

\author{Corina S. P\u{a}s\u{a}reanu}
\email{pcorina@andrew.cmu.edu}
\affiliation{%
 \institution{Carnegie Mellon University, NASA Ames}
  \city{Mountain View}
  \state{California}
  \country{USA}
}


\begin{abstract}
As deep neural networks (DNNs) are increasingly used in safety-critical applications,  there is a growing concern for their reliability. Even highly trained, high-performant networks are not 100\% accurate. However, it is very difficult to predict their behavior during deployment without ground truth. In this paper, we provide a comparative and replicability study on recent approaches that have been proposed to evaluate the reliability of DNNs in deployment. We find that it is hard to run and reproduce the results for these approaches on their replication packages and even more difficult to run them on artifacts other than their own. Further, it is difficult to compare the effectiveness of the approaches, due to the lack of clearly defined evaluation metrics. Our results indicate that more effort is needed in our research community to obtain sound techniques for evaluating the reliability of neural networks in safety-critical domains. To this end, we contribute an evaluation framework that incorporates the considered approaches and enables evaluation on common benchmarks, using common metrics. 

\end{abstract}





\maketitle

\section{Introduction}
Deep Neural Networks (DNNs) have found many applications in safety-critical domains \cite{10.5555/2999325.2999452, 10.1561/0600000079, DBLP:journals/isafm/JorgensenI21} raising important questions about their reliability. Even highly trained, highly accurate DNNs are not 100\% accurate, and can thus contain errors which can have costly consequences in domains such as medical advice \cite{10.5555/2999325.2999452, 10.1561/0600000079}, self-driving cars \cite{10.1561/0600000079}, or banking \cite{DBLP:journals/isafm/JorgensenI21}. However, detecting these errors during deployment is difficult due to the absence of ground truth  at inference time.



In this paper, we aim to provide a comparative and replicability study on recent techniques that were published in the software engineering community that aim to evaluate the reliability of DNNs in deployment. We restrict our attention to white-box, post-hoc methods, that take a pre-trained model, perform a white-box analysis of the internals of the model (possibly guided by labeled data), and produce monitors to be deployed at run time, with the goal of predicting mis-classifications. After a preliminary literature review, we have identified two recent, state-of-the art approaches on this topic: 
SelfChecker~\cite{Xiao2021SelfCheckingDN} and DeepInfer~\cite{ahmed2024inferring}. SelfChecker computes the similarity between DNN layer features of test instances and the samples in the training set, using kernel density estimation (KDE) to detect mis-classifications by the model in deployment. DeepInfer infers data preconditions from the DNN model  and uses them to determine the model's correct or incorrect prediction.
In addition, we propose to use Prophecy \cite{Gopinath2019PropertyIF} for our study. Although it was not used before for determining the reliability of outputs, Prophecy bears similarities to both DeepInfer and SelfChecker, as it also aims to generate data preconditions (as in DeepInfer) albeit at every layer, driven by training data (similar to SelfChecker). 

We first try our best to reproduce the DeepInfer and SelfChecker approaches based on their original artifacts, i.e., tabular datasets and models for DeepInfer vs. image datasets and models for SelfChecker. After that, we go one step further to empirically compare those approaches against each-other's datasets and models. Experimental results indicate that these existing approaches, although very recent and hence well maintained, are difficult to be reproduced on their original artifacts; it is even more difficult to run these approaches on models and datasets other than their own.  Further, for the cases where they can be successfully run on the same artifacts, it is difficult to compare them due to the lack of clear metrics for measuring performance. Our results indicate that significantly more effort should be spent by our community to achieve sound techniques for evaluating the reliability of deep neural networks in deployment. To this end, we propose a common, public framework, TrustDNN, that we built for evaluating and comparing the different techniques, using common metrics. 

The framework incorporates robustified versions of DeepInfer and SelfChecker as well as Prophecy, together with carefully curated datasets and models. Using this framework, we run a comparative evaluation of the three approaches on tabular and image data and respective models. We find that DeepInfer generally performs the best on both tabular data and image data (after our modification). Further, Prophecy is the only approach that can handle both image and tabular data without modifications, while SelfChecker appears to have the highest resource consumption. 
Overall, the results indicate that the three considered approaches have complementary strengths. 
To summarize, we  make the following contributions;
\begin{itemize}
    \item We identify state-of-the-art approaches (SelfChecker~\cite{Xiao2021SelfCheckingDN}, DeepInfer~\cite{ahmed2024inferring}), that evaluate reliability of DNN models and are available as open-source tools. These approaches have commonality in their workflow but are sufficiently different in their methodologies to warrant comparison. 
    \item We present the first-time application of the approach in Prophecy~\cite{Gopinath2019PropertyIF} to evaluate reliability of DNN models.
    \item We develop tools, \textit{TrustBench} and \textit{TrustDNN}, a unified and standard framework to curate benchmarks and tools for replication and comparative studies.
    \item We present a replication study that goes beyond mere re-running of the tools; we (i) enlist all issues encountered and possible solutions, (ii) evaluate existing approaches on new domains, (iii) \textit{refactor and extend} existing approaches to enable re-use and application to new domains.
    \item We present a comparative study of three approaches and discuss observations regarding effectiveness and efficiencies on benchmarks from domains with image and tabular data. 
    \item We have made publicly available the replication package along with artifacts and results (\url{https://zenodo.org/doi/10.5281/zenodo.12803632}). 
\end{itemize}


\section{Approaches} 
\label{sec:approaches}
\begin{figure*}
\centering
\includegraphics[scale=0.35]{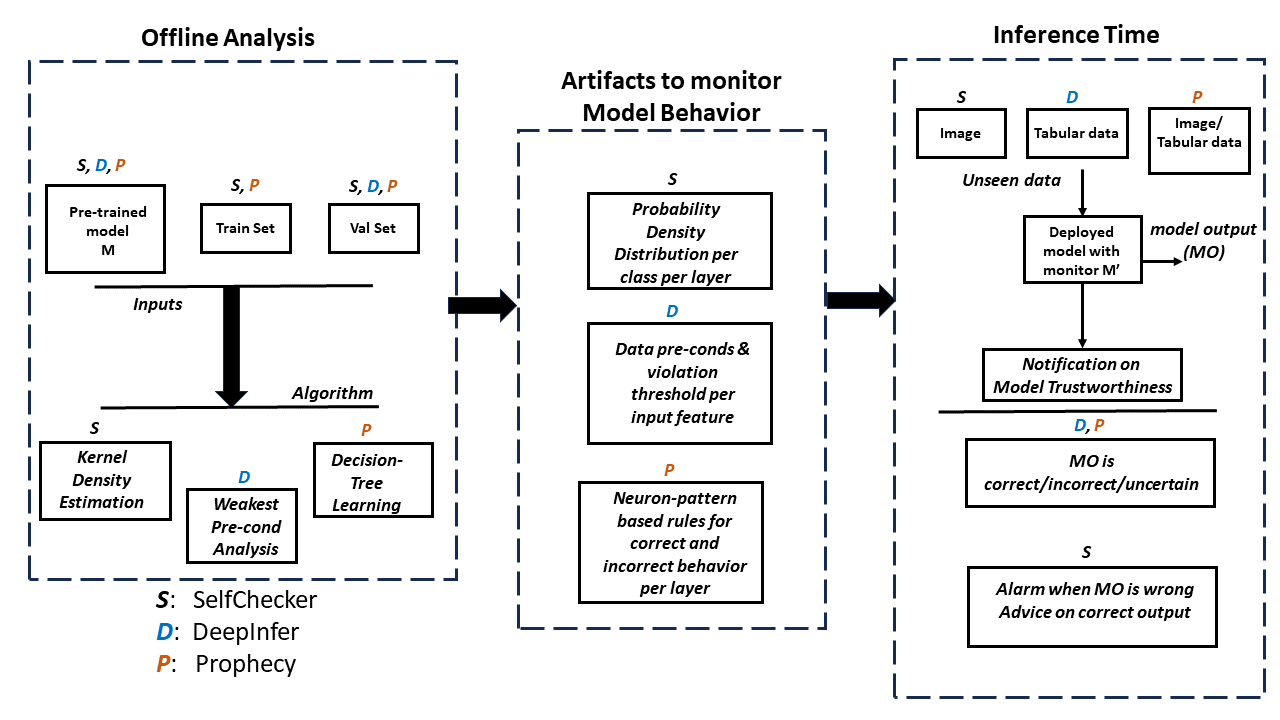}
\caption{Workflow Overview to infer reliability of model behavior.}
\label{fig:workflow}
\end{figure*}




In this section, we briefly survey approaches that measure the reliability of machine learning models. For the purpose of this replicability study, we choose approaches that fit the following criteria;
\begin{itemize}
    \item Techniques that can be considered state-of-the-art: They have been proposed in the past three years or have considerable citations.
    \item Techniques that have been fully developed and made publicly available, and can be quantitatively evaluated.
    \item Approaches that show potential to have broad applicability  (i.e. are not designed to work only for specific input domains).
    \item Techniques whose drawbacks (such as poor accuracy, and poor scalability) have not been highlighted in previous work via thorough evaluation. 
\end{itemize}

Use of confidence scores based on softmax probabilities of DNN classifiers~\cite{HendrycksG16c}, and information theory metrics such as entropy, have been studied by a number of previous work and shown to be unreliable~\cite{8683359},~\cite{NEURIPS2019_757f843a},~\cite{abs-2003-01289}.

The work in ~\cite{jiang2018trust} uses nearest neighbor classifiers to measure model confidence. However, subsequent work ~\cite{NEURIPS2019_757f843a} evaluated this approach to be less efficient and not effective on large datasets and complex models. This work~\cite{NEURIPS2019_757f843a} proposed an approach called ConfidNet, which builds a ConvNet model (on top of the an existing pre-trained model) to learn the confidence criterion based on the true class probability for failure prediction. More recent work ~\cite{Xiao2021SelfCheckingDN} performs a quantitative comparison with ConfidNet to highlight its ineffectiveness when the training data has very few mis-classified instances. Dissector, proposed by Wang et al.~\cite{Wang00M020}, focuses on distinguishing inputs representing unexpected conditions from normal inputs by training a number of sub-models. However, it is applicable mainly for image models. This tool was also evaluated by the more recent work~\cite{Xiao2021SelfCheckingDN}, which highlights that the process of building sub-models is manual and highly time-consuming. SelfOracle ~\cite{0001WCT20} is a technique that monitors video frames or image sequences to identify unsupported driving scenarios based on the estimated model confidence being lesser than a threshold. The approach is specifically designed for self-driving car models and its performance has been evaluated to be poor on other DNN types ~\cite{Xiao2021SelfCheckingDN}. Recent work ~\cite{ahmed2024inferring} further points out that SelfOracle, ConfidNet, and Dissector are not applicable to models processing numeric data.

Based on this overview, we identified two existing approaches for our study; SelfChecker~\cite{Xiao2021SelfCheckingDN}; first proposed in 2021, with more citations than other techniques and with a publicly available tool that was quantitatively compared with previous approaches, and DeepInfer~\cite{ahmed2024inferring}; very recent work due to appear in ICSE 2024 with a publicly available tool. The above two techniques have been individually evaluated only on image (SelfChecker) and tabular data (DeepInfer) respectively but the design is not restrictive to one particular domain. We also include another tool, Prophecy~\cite{Gopinath2019PropertyIF} in our study. This tool  mines assume-guarantee type rules from pre-trained models and we  explore the application of the tool for the first time to determine the reliability of models. We believe it holds potential since the approach is agnostic to the type of inputs.

Figure~\ref{fig:workflow} presents an overview of the workflow of approaches that we identified for determining the reliability of DNN model output at runtime. Every approach typically has an offline analysis phase, which takes as input a pre-trained DNN model, labelled datasets such as the training and validation data, and applies algorithms to capture information learned by the model during training. The output of this phase are artifacts that can be used to monitor model behavior at runtime. The model is accordingly updated with runtime checks before deployment. At inference time, when the model is exposed to inputs (unseen during training), the approaches apply the checks in an attempt to determine if the output produced by the model can be trusted. The output of these approaches is a notification regarding the reliability of the model output and in some cases an advice regarding the correct output. Given that the monitor is built typically based on the information regarding the training profile of the model, they are applicable mainly to predict model behavior of unseen inputs that follow the distribution of the data that the model has been exposed to during training. 

We describe below the three approaches in detail:

\textbf{SelfChecker}: The insight driving this approach is that DNN models typically reach a correct
prediction before the final layer, and in some cases, the final layer may change a correct internal prediction into an incorrect prediction. The features extracted by the internal layers of a model contain information that can be used to check a model’s output. Given a pre-trained model and the training data, the approach applies kernel density estimation (KDE) to determine the probability density distributions (PDD) of each layer’s
output. The PDD corresponding to each class is determined after every layer of the model.

At inference time, for a given input, SelfChecker estimates its probability density for each class within each layer from the computed density functions. The class for which the input instance induces the maximum estimated probability density, is considered as the inferred class for that layer. If a majority of the layers indicate inferred classes that are different from the model prediction, then SelfChecker triggers an alarm. The approach adopts a strategy to select a set of layers that positively contribute to the predication, by evaluating the performance of different combinations of layers on a validation set. A similar technique is adopted to determine the most probable correct prediction and the tool offers this as an advice. It also uses a boosting strategy based on the computed probable correct prediction to increase the alarm accuracy. 

SelfChecker has been mainly evaluated on vision models, and on popular image datasets it has been shown to trigger correct alarms on 60.56\% of wrong DNN predictions, and false alarms on 2.04\% of correct DNN predictions. It has also been evaluated on self-driving car scenarios and has been able to achieve high F1-score (68.07\%) in predicting behavior.

\textbf{DeepInfer}: This work attempts to circumvent the limitation of data-dependency which impacts the effectiveness of approaches such as SelfChecker; the training and validation datasets need to be representative of unseen instances. DeepInfer, on the other hand, applies a pure static analysis based technique, weakest pre-condition analysis using Dijkstra’s Predicate Transformer Semantics on the DNN model to compute pre-conditions on input features. Starting with a post-condition for correct output; DNN output falling within a prediction interval, this analysis is used to compute conditions on the input of the output layer. This process is repeated iteratively until the input layer, to compute  data preconditions. In order to handle multiple layers with hidden non-linearities, this work introduces a novel method for model abstraction and a weakest pre-condition calculus.  

The pre-conditions represent the assumptions made by the pre-trained model on the input profile on which the model can be expected to behave correctly. The pre-conditions for every input variable or feature are evaluated against a validation dataset to obtain a mean value which acts as a threshold. At inference time, the unseen input is checked against the data pre-conditions and the violations are compared with the respective thresholds to predict if the model output is correct or incorrect. If sufficient evidence is not available to make a concrete decision, an "uncertain" output is produced.

The applicability, effectiveness and scalability of this approach is dependent on the architecture of the model, the precision of the abstraction and the input dimensionality. Unlike previous work, this technique has not been evaluated on vision models. It has been shown to perform well on multiple models with tabular data whose input dimensionality is much lower than images. 

\textbf{Prophecy}: Given a DNN model $F$ and an output property $P(F(X))$, Prophecy~\cite{Gopinath2019PropertyIF} extracts rules of the form, $\forall X: \sigma(X) \Rightarrow P(F(X))$, which can be viewed as {\it abstractions of model behavior}. $\sigma(X)$ is a pre-condition in terms of neuron patterns at the inner layers of the network; constraints on the values or activations ({\em on}, {\em off}) of a subset of neurons at one or more layers. $P(F(X))$ could be any property encoded as a constraint on the model output such as; a classifier's output being a certain label, the label being equal to ideal (correctness), the outputs of a regression model being within a certain range (safety) so on.

Previous work (~\cite{Gopinath2019PropertyIF},~\cite{KimGPS20},~\cite{KadronGPY21},~\cite{UsmanGSP22},~\cite{UsmanGSNP21},~\cite{Usman0024SGP23}), has explored the use of Prophecy in obtaining formal guarantees of behavior, repair, obtaining explanations, debugging and testing. We see potential in using Prophecy to determine model reliability, which we explore in this work. The neuron-patterns based rules at a given layer of the model capture the logic (in terms of internal layer features). Given a set of passing and failing data, Prophecy could be used to mine rules corresponding to correct and incorrect model outputs, which in turn potentially capture the correct and vulnerable portions of the model respectively. Each rule has a formal mathematical form, which enables a quick and a precise evaluation on inputs. Further, the rules correspond to the logic learned by the model during training, irrespective of the nature or dimensionality of the input (image,tabular,text so on). Vision models typically comprise of an encoder portion which extracts features from the raw images and a head that uses the extracted features to make decisions. We envisage applying Prophecy to the dense layers following the encoder to capture this logic. 

The offline analysis phase comprises of feeding Prophecy with a pre-trained model and a dataset labelled with passing and failing inputs. Prophecy uses decision-tree learning to extract rules corresponding to correct and incorrect model outputs in terms of the neuron values at every layer. At inference time, given an unseen input, it is evaluated against the set of rules for correct and incorrect output respectively at every layer. The rules at a given layer are mutually exclusive, therefore, the input either satisfies one of the rules for correct behavior, or one of the rules for incorrect behavior, and this corresponds to the decision as predicted by the rules at the layer. The correct/incorrect decision corresponds to the one that receives majority of votes. If there is a tie, the output is an "uncertain" decision.

\subsection{Challenges}
\label{sec:challenges}
During our preliminary review of the 
approaches, we identified the following challenges for our study: 

\begin{itemize}
    \item \textit{Lack of commonality in the input domains:} Existing approaches operate and have been evaluated on either image data (SelfChecker), or non-image data such as tabular data (DeepInfer). There is no common set of benchmarks (from different domains) on which tools measuring reliability could be evaluated on. 
    
    \item \textit{Ad-hoc and non-uniform data preparation methods:}  We found that each tool adopts a different technique to obtain and process the data (train, test, validation datasets and models). It is essential to have a standard way to obtain the raw data and models, pre-process them and curate them for use. 
    
    \item \textit{Problems with setup:} We found a couple of issues with how the approaches are implemented, which impedes ease of use and generalizability. Many parameters are hard-coded; eg. the layers to be considered for a given model, the use of pre-computed values, and so on. The replication package of DeepInfer has different scripts for the same function for different models, has multiple folders containing duplicate information (models, data), which causes confusion and are also error-prone.

    \item \textit{Lack of proper documentation:} There is very little documentation on the tool websites which clarifies the data preparation methods and the setup.
    
    \item \textit{Lack of standard metrics for evaluation:}
    Discrepancies in evaluation methodologies and metrics across approaches complicate fair and compatible comparisons. Both the existing approaches we consider, make use of the confusion matrix metrics, namely true positive (TP), false positive (FP), true negative (TN), and false negative (FN). They use True Positive Rate (TPR), False Positive Rate (FPR), and F1-score to evaluate their techniques and compare them with others. However, there are differences in the way these metrics are actually computed in the respective codes, which makes it difficult to interpret if they have the same meaning. Further, the code that computes these metrics is closely coupled with the code implementing the functionality of the approach, which further hinders interpretation and also inhibits the application of other metrics to evaluate them.

    \item \textit{Inherent difficulty in evaluating machine learning models:} Machine learning is a highly data-dependent process. Most of the approaches that are used to evaluate reliability of other models, themselves use data-driven algorithms. Therefore, in order to replicate results, we need the exact data used by the approaches in their analysis and inference phases. 
    The inherent randomness in the training process of models can lead to varied outcomes. So even if we have access to the exact data, the results may still be different. 
\end{itemize}




\section{Methods}
\label{sec:methods}

\begin{figure}[htbp]
    \centering
    \includegraphics[scale=0.35]{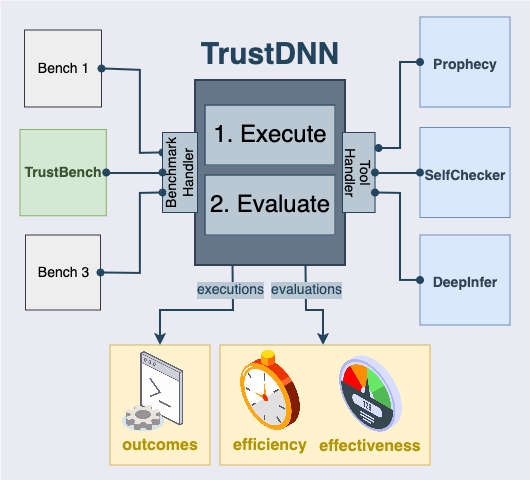}
    \caption{The TrustDNN Framework.}
    \label{fig:framework}
\end{figure}

This section describes our methodology in performing this replicability and comparative study. We developed tools, namely \textit{TrustBench} and \textit{TrustDNN}, which seek to address the aforementioned challenges.

\subsection{TrustBench}
\label{sec:trustbench}

We designed TrustBench with Gray's definition of a good benchmark~\cite{gray1991benchmark}, considering it is repeatable, portable, scalable, representative, requires minimum changes in the target tools, and is simple to use. TrustBench is repeatable by automating the collection and preparation of datasets and models from different domains in a standard manner. TrustBench ensures portability by using JSON files to define datasets, models, and other configurations, making it easy to share and replicate the setup across different systems. Additionally, its integration with the APIs of popular data sources like Kaggle~\footnote{\url{https://www.kaggle.com/} - the largest AI and ML community.} and Keras~\footnote{Keras - \url{https://keras.io/}} enables it to fetch and use data consistently across various environments. TrustBench has a straightforward Command Line Interface (CLI) that features only three commands (collect, prepare, detail), making it easy to use. We made TrustBench representative regarding two previous tools~\cite{Xiao2021SelfCheckingDN, ahmed2024inferring} by including their curated version of datasets and models. We present a detailed description of the process followed and issues encountered in section~\ref{subsec:dataprep}. TrustBench is scalable by allowing easy integration of new data sources just by extending a class; refer to our public repository~\footnote{ \url{https://github.com/epicosy/TrustBench}} for detailed instructions. Lastly, TrustBench does not require changes in the target tools as it outputs artifacts for the datasets, models, and predictions at fine granularity and in a structured hierarchy.

\subsection{TrustDNN Framework}
\label{sec:framework}
A good framework is simple and quick to use, reusable in design and code, and provides enough adjustable and extensible features~\cite{Roberts2004EvolvingFA}. To compare various approaches in a unified set-up, we have developed the TrustDNN framework, illustrated in Figure~\ref{fig:framework}. TrustDNN was built using Cement~\footnote{\url{https://builtoncement.com/} - CLI framework for Python}, a CLI framework that provides a robust interface/handler system. Leveraging that, TrustDNN offers a plugin system with dedicated handlers: a \textit{ToolHandler} and \textit{BenchmarkHandler}. These handlers facilitate the integration of new tools and benchmarks as plugins, each configurable through JSON files. This configurable and modular plugin system makes TrustDNN adjustable, extensible, and reusable in design and code. The framework follows a two-step workflow (\textit{execute} and \textit{evaluate}), delivered through two commands, each with two actions, which makes it simple and quick to use. In the \textit{execute} step, the \textit{analysis} and \textit{inference} phases are executed separately for each tool. The tools are expected to produce a file containing notifications as output after execution. Subsequently, during evaluation, the \textit{efficiency} and \textit{effectiveness} of the tools are measured by analyzing successful executions and their respective outputs. We provide more details on the usage of TrustDNN in our public repository~\footnote{\url{https://github.com/epicosy/TrustDNN}}.

The efficiency is typically measured in terms of time and memory consumption. To measure time efficiency, we register the duration of the process executing the tool. For memory consumption, we continuously monitor the Resident Set Size (RSS) of the process and its children, summed at intervals of 200 milliseconds. Following execution, we record the peak RSS. However, it's important to note that RSS only accounts for non-swapped physical memory usage and may not fully reflect all memory types, such as virtual memory. 

The effectiveness of any algorithm making predictions, such as estimating the reliability of other models, measures how close the predictions are to the actual. We describe in detail the metrics used for this study in section~\ref{subsec:metrics}.


\subsection{TrustBench and TrustDNN for our study}
\label{sec:Trust_our_study}
In this section, we highlight the details specific to applying TrustBench and TrustDNN for this study.

\subsubsection{\bf Data preparation using TrustBench}
\label{subsec:dataprep}

The replication package of DeepInfer~\cite{DeepInferRepo} provides artifacts for four case studies for tabular data, namely \textit{Bank Marketing (BM)}, \textit{German Credit (GC)},\textit{ House Price (HP)}, and \textit{PIMA Diabetes (PD)}. There are totally 29 models corresponding to these case-studies. The replication package contains a Data folder with a single dataset for each case-study, representing the unseen or test data. A standard benchmark dataset for machine learning models and tools consists of train, validation and test data. Therefore, we employed TrustBench to prepare these datasets for the 4 case studies. After consultation with the authors of ~\cite{ahmed2024inferring}, we obtained the respective datasets from Kaggle and performed a standard train/validation/test split, with proportions specified by Xiao et al.~\cite{Xiao2021SelfCheckingDN} – allocating 80\% for training and 10\% each for validation and testing. 

\textbf{Bank Marketing (BM)}~\cite{BankMarketingDataset}: To prepare this dataset, we referred to the code provided in the replication package~\cite{DeepInferRepo} and
consulted the most popular notebook associated with it on Kaggle~\cite{BankMarketingPreprocessing}, which follows a similar pre-processing approach. We followed the same steps by balancing each class to contain 5289 samples. However, this method is tailored to this binary scenario and may not be suitable for broader applications. Next, we employed categorical encoding for the columns representing \textit{month}, \textit{education}, and \textit{outcome}. Subsequently, we utilized One-Hot Encoding via \textit{get\_dummies} function in Pandas, but this time individually for the remaining categorical fields (\textit{job}, \textit{marital}, \textit{default}, \textit{housing}, \textit{loan}, \textit{contact}) rather than applying it to the entire dataframe. Notably, a discrepancy arose in the one-hot encoded columns. Specifically, we obtained additional columns such as \textit{job\_admin}, \textit{marital\_divorced}, \textit{default\_no}, \textit{housing\_no}, \textit{loan\_no}, and \textit{contact\_cellular}. This inconsistency contradicts the expected behavior outlined in the documentation for the \textit{get\_dummies} function, which specifies the conversion of each variable into binary (0/1) variables corresponding to its unique values. Given this concern, we excluded those additional columns from our data preparation process to maintain compatibility with existing pre-trained models.

\textbf{German Credit (GC)}~\cite{GermanCreditDataset}: 
For preparing this dataset, we consulted the code provided in the replication package~\cite{DeepInferRepo}
and the notebook associated with the GC dataset that follows the most similar preprocessing steps~\cite{GermanCreditKaggle}. Following the same approach, we normalized numerical columns to ensure a uniform order of magnitude and applied one-hot encoding to categorical columns. For normalization, we utilized the min-max scaler from \textit{scikit-learn}\footnote{\url{https://scikit-learn.org/stable/} - Python package for ML and data analysis} to scale the data between 0 and 1. The one-hot encoding process mirrored that used for the BM dataset. Additionally, the \textit{Risk} column was transformed into a binary variable since it serves as the classification label. Interestingly, we observed in the GC preprocessing script within the replication package that the column \textit{Saving\ accounts\_rich} was dropped without explanation. We do the same, and following data preparation, we observe one additional column, \textit{Purpose\_vacation/others}, which we again excluded to maintain compatibility with existing pre-trained models.

\textbf{House Price (HP)}~\cite{HousePriceDataset}: For this dataset, we consulted the script in the replication package~\cite{DeepInferRepo} and the notebook associated with it on kaggle~\cite{HousePriceKaggle} to apply the same min-max scaling as in the preparation of the GC dataset. 

\textbf{PIMA Diabetes (PD)}~\cite{PIMADataset}: For preparing this dataset, we performed our train/val/test split with the same random state 10 considered in the script provided in the replication package~\cite{DeepInferRepo}.

\midsepremove
\begin{table}[htbp]
{\footnotesize
    \centering
    \caption{Description of datasets (\textit{\#L} refers to labels count).}
    \label{table:datasets}
    \begin{subtable}{\linewidth}
        \centering
        \begin{tabular}{c | r c c r | r r}
            \toprule
            \headcol
             & \multicolumn{4}{c|}{\textbf{\#Samples}} & \textbf{Input} & \\
            \cmidrule{2-5}
            \headcol
            \multirow{-2}{*}{\textbf{Dataset}} & \textit{Total} & \textit{Train} & \textit{Val.} & \textit{Test} & \textbf{Size} & \multirow{-2}{*}{\textbf{\#L}} \\
            \midrule
            BM~\cite{DeepInferRepo} & 2 116 & NA & NA & 2 116 & 28 & 2 \\
            \rowcol
            GC~\cite{DeepInferRepo} & 200 & NA & NA & 200 & 22 & 2 \\
            HP~\cite{DeepInferRepo} & 292 & NA & NA & 292 & 10 & 2 \\
            \rowcol
            PD~\cite{DeepInferRepo} & 153 & NA & NA & 153 & 8 & 2\\
            \bottomrule
        \end{tabular}
        \caption{Tabular datasets: Original artifacts provided by DeepInfer~\cite{DeepInferRepo}}
        \label{tab:tableA}
    \end{subtable}

    \begin{subtable}{\linewidth}
        \centering
        \begin{tabular}{c | r r r r | r r}
            \toprule
            \headcol
             & \multicolumn{4}{c|}{\textbf{\#Samples}} & \textbf{Input} & \\
            \cmidrule{2-5}
            \headcol
            \multirow{-2}{*}{\textbf{Dataset}} & \textit{Total} & \textit{Train} & \textit{Val.} & \textit{Test} & \textbf{Size} & \multirow{-2}{*}{\textbf{\#L}} \\
            \midrule
            BM~\cite{BankMarketingDataset} & 10 578 & 8 462 & 1 058 & 1 058 & 28 & 2 \\
            \rowcol
            GC~\cite{GermanCreditDataset} & 1 000 & 800 & 100 & 100 & 22 & 2 \\
            HP~\cite{HousePriceDataset} & 1 460 & 1 168 & 146 & 146 & 10 & 2 \\
            \rowcol
            PD~\cite{PIMADataset} & 768 & 614 & 77 & 77 & 8 & 2\\
            \bottomrule
        \end{tabular}
        \caption{Tabular datasets: Prepared using TrustBench.}
        \label{tab:tableB}
    \end{subtable}%

    \begin{subtable}{\linewidth}
        \centering
        \begin{tabular}{c | r r r r | r r}
            \toprule
            \headcol
             & \multicolumn{4}{c|}{\textbf{\#Samples}} & \textbf{Input} & \\
            \cmidrule{2-5}
            \headcol
            \multirow{-2}{*}{\textbf{Dataset}} & \textit{Total} & \textit{Train} & \textit{Val.} & \textit{Test} & \textbf{Size} & \multirow{-2}{*}{\textbf{\#L}} \\
            \midrule
            CIFAR~\cite{CIFAR10Dataset} & 60 000 & 40 000 & 10 000 & 10 000 & 32*32*3 & 10 \\
            \bottomrule
        \end{tabular}
        \caption{Standard Image dataset used by SelfChecker.}
        \label{tab:tableC}
    \end{subtable}%
    }
\end{table}
\midsepdefault

Table~\ref{tab:tableA} contains details on the datasets in the original replication package of DeepInfer. 
Table~\ref{tab:tableB} contains details on the datasets prepared using TrustBench for reuse by other approaches.

The replication package of SelfChecker~\cite{SCRepo} contains artifacts corresponding to only \textbf{CIFAR10}. This is a standard and popular benchmark ~\cite{CIFAR10Dataset} 
, which we obtain from Keras~\cite{CIFAR10Keras} and for which we follow the exact data preparation as performed by Y. Xiao et al. in their replication package ~\cite{SCRepo}.


\subsubsection{\bf Models}
\midsepremove
\begin{table}[htbp]
{\footnotesize
    \centering
    \caption{Description of the models.}
    \label{table:benchmark}
    \begin{tabular}{c | c | r r | r r r}
        \toprule
        \headcol
         &  & \multicolumn{2}{c|}{\textbf{GT}} & & & \\
        \cmidrule{3-4}
        \headcol
        \multirow{-2}{*}{\textbf{Dataset}} & \multirow{-2}{*}{\textbf{Model}} & \textit{\#C} & \textit{\#I} & \multirow{-2}{*}{\textbf{\#L}} & \multirow{-2}{*}{\textbf{\#P}} & \multirow{-2}{*}{\textbf{Acc.}} \\
        \midrule
        \cellcolor{lavender} & BM1 & 856 & 202 & 3 & 2 913 & 80.91\% \\
        \rowcol
         & BM2 & 863 & 195 & 3 & 1 473 & 81.57\% \\
         \cellcolor{lavender} & BM3 & 850 & 208 & 2 & 3001 & 80.34\% \\
        \rowcol
         & BM4 & 846 & 212 & 4 & 24551 & 79.96\% \\
         \cellcolor{lavender} & BM5 & 856 & 202 & 3 & 879 & 80.91\% \\
         \rowcol
         \textit{Bank} & BM6 & 854 & 204 & 3 & 361 & 80.72\% \\
         \textit{Customers} \cellcolor{lavender} & BM7 & 861 & 197 & 3 & 6081 & 81.38\% \\
         \rowcol
         & BM8 & 868 & 190 & 6 & 4641 & 82.04\% \\
         \cellcolor{lavender} & BM9 & 857 & 201 & 3 & 627 & 81.0\% \\
         \rowcol
         & BM10 & 520 & 538 & 3 & 627 & 49.15\% \\
         \cellcolor{lavender} & BM11 & 520 & 538 & 3 & 627 & 49.15\% \\
         \rowcol
         & BM12 & 869 & 189 & 4 & 1439 & 82.14\% \\
        \midrule
         \cellcolor{lavender} & GC1 & 67 & 33 & 2 & 1201 & 67.0\% \\
         \rowcol
         & GC2 & 67 & 33 & 2 & 2401 & 67.0\% \\
         \cellcolor{lavender} & GC3 & 67 & 33 & 2 & 217 & 67.0\% \\
         \rowcol
         & GC4 & 67 & 33 & 3 & 171 & 67.0\% \\
         \textit{German} \cellcolor{lavender} & GC5 & 67 & 33 & 6 & 4257 & 67.0\% \\
         \rowcol
         \textit{Credit} & GC6 & 67 & 33 & 3 & 2295 & 67.0\% \\
         \cellcolor{lavender} & GC7 & 67 & 33 & 3 & 2295 & 67.0\% \\
         \rowcol
         & GC8 & 67 & 33 & 3 & 2295 & 67.0\% \\
        \cellcolor{lavender} & GC9 & 67 & 33 & 4 & 2801 & 67.0\% \\
        \midrule
        \cellcolor{lavender} & HP1 & 124 & 22 & 3 & 273 & 84.93\% \\
        \rowcol
        \textit{House} & HP2 & 122 & 24 & 3 & 273 & 83.56\% \\
        \cellcolor{lavender} \textit{Price} & HP3 & 74 & 72 & 3 & 273 & 50.68\% \\
        \rowcol
         & HP4 & 128 & 18 & 4 & 383 & 87.67\% \\
         \midrule
        \cellcolor{lavender} & PD1 & 61 & 16 & 3 & 221 & 79.22\% \\
        \rowcol
        \textit{Pima} & PD2 & 45 & 32 & 3 & 221 & 58.44\% \\
        \cellcolor{lavender} \textit{Diabetes} & PD3 & 46 & 31 & 3 & 221 & 59.74\% \\
        \rowcol
         & PD4 & 62 & 15 & 4 & 293 & 80.52\% \\
        \midrule
        \cellcolor{lavender} \textit{CIFAR10} & ConvNet & 8045 & 1955 & 25 & 2.9M & 80.45\% \\
        \bottomrule
        \multicolumn{7}{c}{\textit{\#C} - Num. of Correct; \textit{\#I} - Num. of Incorrect;}\\
        \multicolumn{7}{c}{\textit{\#L} - Num. of Layers; \textit{\#P} - Num. of Trainable Parameters;}
    \end{tabular}
    }
\end{table}
\midsepdefault

We use the 29 models provided in the replication package from \cite{ahmed2024inferring} for tabular data. For the image data, we use the only model (ConvNet) provided in the replication package for SelfChecker~\cite{Xiao2021SelfCheckingDN}. Table~\ref{table:benchmark} details the characteristics of the models used in this study and their accuracy on the respective test sets.





\subsubsection{\bf Tools}
We examined the repositories for the tools, DeepInfer and SelfChecker, in order to integrate them into the TrustDNN framework. We found several problems as highlighted in the challenges; the repository for DeepInfer contains scripts specialized for each model and dataset and has un-necessary replication of code. We found SelfChecker's implementation to be tightly coupled with the computation of the evaluation metrics, which impedes obtaining the exact output notifications from the tool. 

We refactored the code for each approach , based on consultation with the respective authors, to make it easy to apply to new models and datasets. We removed duplication, made the code more modular and generalizable by decoupling the implementation from specific datasets, models, and evaluation metrics. However, we were unable to refactor SelfChecker's code to obtain the exact output notifications for a given test data, instead of the overall evaluation metrics. We have communicated this issue to the authors.


In order to use Prophecy for the purpose of this study, as explained in Section~\ref{sec:approaches}, we used it to build layer-wise pre-conditions to capture the decision logic of the model in terms of features learnt at the layer. Specifically, Prophecy builds a decision-tree estimator in terms of neuron values at every layer, to predict correct and incorrect model behavior. We implemented an efficient algorithm that directly invokes this estimator at inference time for the purpose of runtime detection.

A noteworthy point is that the layers of a model's architecture that are considered by each tool is driven by the design rationale of the approach. DeepInfer is implemented to consider all layers of the model, in order to build pre-conditions at the input layer. SelfChecker, on the other hand, considers only the activation layers to build PDDs based on layer features after each such layer. Prophecy works on the dense and activation layers; it considers only the layers of the classification head for vision models 
TrustDNN provides the ability to set suitable configuration parameters to enable the layer selection for each tool.
\subsubsection{\bf Metrics}
\label{subsec:metrics}
SelfChecker and DeepInfer use similar metrics but we found discrepancies, as pointed out in the challenges (Section~\ref{sec:challenges}). 
Specifically, there is ambiguity in defining the {\em positive} and {\em negative} classes when using True Positives (TP), False Positives (FP), False Negatives (FN), and True Negatives (TN) metrics. Traditionally, the {\em positive} class signifies correct predictions, while the {\em negative} class signifies incorrect ones. However, SelfChecker flips this association, considering incorrect predictions as {\em positive} and correct ones as {\em negative}. Although unconventional, we adopt SelfChecker's approach in our paper as it provides a meaningful evaluation of a technique's performance with respect to its ability to capture misclassifications accurately. 


In this study, we propose to use the Matthews Correlation Coefficient (MCC) as our primary metric for comparing the effectiveness of the approaches for several reasons. First, MCC considers all four categories of the confusion matrix, making it immune to class swapping and comprehensively assessing the performance of a tool~\cite{Chicco2020TheAO}. This ensures a fair comparison between tools, regardless of how the positive and negative classes are defined. This is clearer when we observe the discrepancy in reported metrics between DeepInfer and SelfChecker. While DeepInfer claims alignment with Self-Checker's metrics, there are apparent inconsistencies in their reported numbers on true positives and false positives, raising concerns about relying solely on traditional metrics. By incorporating MCC, we enhance the reliability and interpretability of our evaluation, mitigating the risk of misinterpretations or discrepancies in reported metrics. Furthermore, MCC considers the distribution of positive and negative elements in the dataset, offering a balanced evaluation of the capability of a tool to classify instances across all classes. This balanced assessment is crucial, especially in detecting misclassified inputs that are intrinsically less frequent.

MCC represents the correlation between the predicted and actual classifications, with values ranging from -1 (total disagreement) to +1 (perfect agreement), where 0 indicates random classification. The MCC calculation is based on the differences between the observed and expected classifications, considering both the overall agreement and the balance between the positive and negative classifications.

\begin{equation}
\text{MCC} = \frac{\text{TP} \times \text{TN} - \text{FP} \times \text{FN}}{\sqrt{(\text{TP} + \text{FP})(\text{TP} + \text{FN})(\text{TN} + \text{FP})(\text{TN} + \text{FN})}}
\end{equation}

To provide a more comprehensive overview of our results, we incorporate metrics such as True Positive Rate (TPR), False Positive Rate (FPR), and F1 Score alongside MCC. While TPR, FPR, and F1 Score may bias the assessment towards the positive class and overlook certain aspects of the confusion matrix, they offer valuable insights when appropriately interpreted. Further, these metrics were used in the previous studies on DeepInfer, SelfChecker, and Prophecy.\\

\begin{itemize}
    \item{\textbf{False Positive Rate (FPR)} quantifies the tendency of a tool to misclassify negative instances as positive with the following equation: 

    \begin{equation}
    \text{FPR} = \frac{\text{False Positives}}{\text{False Positives} + \text{True Negatives}}
    \end{equation}
    
    FPR evaluation is crucial for tasks prioritizing the reduction of false alarms.
    }\\
    \item{\textbf{True Positive Rate (TPR)}, also known as sensitivity or recall, measures the ability of a tool to identify positive instances correctly and is computed as:

    \begin{equation}
    \text{TPR (Recall)} = \frac{\text{True Positives}}{\text{True Positives} + \text{False Negatives}}
    \end{equation}
    
    TPR evaluation is essential for assessing positive instance detection, especially in scenarios where capturing all positive cases is critical.}\\
    \item{\textbf{Precision} assesses the reliability of positive predictions made by the tool, and its usage is particularly relevant in contexts where minimizing false positives is paramount. It is computed with the following:

    \begin{equation}
    \text{Precision} = \frac{\text{True Positives}}{\text{True Positives} + \text{False Positives}}
    \end{equation}
    }\\
    \item{\textbf{F1 Score} provides a balanced assessment by considering the harmonic mean of precision and recall with the following: 

    \begin{equation}
    \text{F1 Score} = 2 \times \frac{\text{Precision} \times \text{Recall}}{\text{Precision} + \text{Recall}}
    \end{equation}
    
    F1 Score offers a comprehensive view of the overall performance of a tool and is useful in situations characterized by imbalanced positive and negative instances, as it accounts for false positives and negatives.
    }\\
\end{itemize}

We would like to draw attention to the fact that the \textit{uncertains} are considered as incorrect in the metric formulae.

\section{Evaluation}\label{sec:results}


We aim to address the following \textbf{research questions} in this study.

\begin{itemize}
\item RQ1: For DeepInfer and SelfChecker, to what extent can the results be reproduced on their own artifacts (datasets and models)? 
\item RQ2: For the considered approaches,  can the results be obtained on artifacts other than their own?
\item RQ3: How do the considered approaches compare in terms of effectiveness?
\item RQ4: How do the considered approaches compare in terms of time and memory consumption?
\end{itemize}

\subsection{Experimental Design and Setup}

We ran two sets of experiments: {\em replicability analysis} (to answer RQ1 and RQ2) and {\em comparative analysis} (to answer RQ3 and RQ4).

For RQ1 in the {\em replicability analysis}, we aim to run each tool 
on the exact configuration, models, and datasets provided by the respective replication packages. 

To address RQ2 and for the {\em comparative analysis}, we execute each tool using the datasets and models prepared by TrustBench, as described in Section~\ref{sec:trustbench}.  
Listings 1 and 2 in appendix specify the exact configuration parameters set for each tool in TrustDNN. 

The experiments were executed on a Debian-based system with the Linux kernel version 5.10.0-16-amd64 running on a remote server with 128 GB of RAM and a 2.1 GHz Intel(R) Xeon(R) Silver 4130 with 48 cores.

\subsection{Results}

\midsepremove
\begin{table}[!ht]
{\footnotesize
    \centering
    \caption{Replication results for DeepInfer}
    \label{table:rq1}
    \begin{tabular}{|c|r|r|r|r|r|r|r|}
    \hline
        \headcol
         &  \multicolumn{2}{c|}{\textbf{Ground Truth}} & \multicolumn{5}{c|}{\textbf{DeepInfer}} \\
        \cmidrule{2-3}
        \cmidrule{4-8}
        \headcol
         \multirow{-2}{*}{\textbf{Model}} & \textit{\#C} & \textit{\#I} & \textit{\#C} & \textit{\#I} & \textit{\#U} & \textit{\#Violation} & \textit{\#Satisfaction} \\
        \toprule
        \mintrow
        \cellcolor{lavender} PD1 & 119 & 34 & 108 & 43 & 2 & 192 & 1032 \\
        \mintrow
        \cellcolor{lavender} PD2 & 99 & 54 & 153 & 0 & 0 & 0 & 1224 \\ 
        \mintrow
        \cellcolor{lavender} PD3 & 98 & 55 & 74 & 79 & 0 & 129 & 1095\\ 
        \mintrow
        \cellcolor{lavender} PD4 & 111 & 42 & 37 & 116 & 0 & 132 & 1092 \\ 
        \midrule
        \mintrow
        \cellcolor{lavender} HP1 & 147 & 145 & 188 & 98 & 6 & 341 & 2579 \\
        \mintrow
        \cellcolor{lavender} HP2 & 147 & 145 & 188 & 98 & 6 & 341 & 2579 \\ 
        \mintrow
        \cellcolor{lavender} HP3 & 145 & 147 & 292 & 0 & 0 & 0 & 2920 \\ 
        \mintrow
        \cellcolor{lavender} HP4 & 147 & 145 & 107 & 184 & 1 & 188 & 2732 \\ 
        \midrule
        \cellcolor{lavender} BM1 & 1713 & 403 & \cellcolor{mint} 616 & \cellcolor{mint} 1500 & \cellcolor{mint} 0 & \cellcolor{mint} 18814 & \cellcolor{mint} 40434 \\ 
        \rowcol
        BM2 & 1724 & 392 & \cellcolor{mint} 1492 & \cellcolor{mint} 624 & \cellcolor{mint} 0 & \cellcolor{mint} 14855 & \cellcolor{mint} 44393 \\ 
        \cellcolor{lavender} BM3 & 1707 & 409 & \cellcolor{mint} 734 & \cellcolor{mint} 1382 & \cellcolor{mint} 0 & \cellcolor{mint} 7370 & \cellcolor{mint} 51878 \\ 
        \rowcol
        BM4 & 1663 & 453 & \cellcolor{mint} 1061 & \cellcolor{mint} 1055 & \cellcolor{mint} 0 & \cellcolor{mint} 17486 & \cellcolor{mint} 41762 \\
        \cellcolor{lavender} BM5 & 1734 & 382 & 998 & 1118 & 0 & 11970 & 47278 \\ 
        \rowcol
        BM6 & 1717 & 399 & 609 & 1507 & \cellcolor{mint} 0 & 18874 & 40374 \\ 
        \cellcolor{lavender} BM7 & 1721 & 395 & \cellcolor{mint} 1375 & \cellcolor{mint} 741 & \cellcolor{mint} 0 & \cellcolor{mint} 12762 & \cellcolor{mint} 46486 \\ 
        \rowcol
        BM8 & 1732 & 384 & \cellcolor{mint} 1089 & \cellcolor{mint} 1027 & \cellcolor{mint} 0 & \cellcolor{mint} 15213 & \cellcolor{mint} 44035 \\
        \cellcolor{lavender} BM9 & 1723 & 393 & 2116 & 0 & \cellcolor{mint} 0 & 17090 & 42158 \\ 
        \rowcol
        BM10 & \cellcolor{mint} 1092 & \cellcolor{mint} 1024 & 1057 & 1059 & \cellcolor{mint} 0 & 13972 & 45276 \\ 
        \cellcolor{lavender} BM11 & \cellcolor{mint} 1092 & \cellcolor{mint} 1024 & 863 & 1253 & \cellcolor{mint} 0 & 18353 & 40895 \\ 
        \rowcol
        BM12 & 1727 & 389 & 1072 & 1044 & \cellcolor{mint} 0 & 17942 & 41306 \\ 
        \midrule
        \cellcolor{lavender} GC1 & \cellcolor{mint} 198 & \cellcolor{mint} 2 & 111 & 89 & \cellcolor{mint} 0 & \cellcolor{mint} 1044 & \cellcolor{mint} 3356\\
        \rowcol
        GC2 & \cellcolor{mint} 198 & \cellcolor{mint} 2 & 48 & 152 & \cellcolor{mint} 0 & \cellcolor{mint} 959 & \cellcolor{mint} 3441 \\
        \cellcolor{lavender} GC3 & \cellcolor{mint} 198 & \cellcolor{mint} 2 & 150 & 50 & \cellcolor{mint} 0 & 1417 & 2983 \\
        \rowcol
        GC4 & \cellcolor{mint} 198 & \cellcolor{mint} 2 & 113 & 87 & \cellcolor{mint} 0 & 2469 & 1931\\
        \cellcolor{lavender} GC5 & \cellcolor{mint} 198 & \cellcolor{mint} 2 & 63 & 137 & \cellcolor{mint} 0 & \cellcolor{mint} 1193 & \cellcolor{mint} 3207 \\
        \rowcol
        GC6 & \cellcolor{mint} 198 & \cellcolor{mint} 2 & 137 & 63 & \cellcolor{mint} 0 & 1328 & 3072\\
        \cellcolor{lavender} GC7 & \cellcolor{mint} 198 & \cellcolor{mint} 2 & 57 & 143 & \cellcolor{mint} 0 & 979 & 3421\\
        \rowcol
        GC8 & \cellcolor{mint} 198 & \cellcolor{mint} 2 & 135 & 65 & \cellcolor{mint} 0 & 1507 & 2893 \\
        \cellcolor{lavender} GC9 & \cellcolor{mint} 198 & \cellcolor{mint} 2 & 76 & 124 & \cellcolor{mint} 0 & 1580 & 2820\\
        \bottomrule
        \multicolumn{8}{c}{\textit{\#C} - Correct; \textit{\#I} - Incorrect; \textit{\#U} - Uncertain;}\\
    \end{tabular}
    }
\end{table}
\midsepdefault

\midsepremove
\begin{table}[htbp]
{\footnotesize
    \centering
    \caption{Replication results for SelfChecker.}
    \label{table:selfchecker}
    \begin{tabular}{|c | r r r r | r r r|}
    \toprule
    \headcol
    & \multicolumn{4}{c|}{\textbf{Confusion Matrix}} & \multicolumn{3}{c|}{\textbf{Metrics}} \\
    \cmidrule{2-5}
    \cmidrule{6-8}
    \headcol
    \multirow{-2}{*}{\textbf{Model}} & \textit{TP} & \textit{FP} & \textit{TN} & \textit{FN} & \textit{TPR} & \textit{FPR} & \textit{F1} \\
    \midrule
    \cellcolor{lavender} \textit{CIFAR10} & 1251 & 239 & 7806 & 704 & 63.99\% & 2.97\% & 72.63\% \\
    \rowcol
    \bottomrule
    \end{tabular}
    }
\end{table}
\midsepdefault


{\bf RQ.1: For the considered approaches, to what extent can the results be reproduced on their own artifacts?}

As Prophecy was not considered before for the problem at hand we do not have a replication package for it. So for this question we only consider SelfChecker and DeepInfer which both come with replication packages. We evaluate DeepInfer on its original datasets (Table~\ref{tab:tableA})  and pre-trained models and SelfChecker with the image dataset (Table~\ref{tab:tableC}) and model. 

The paper describing DeepInfer \cite{ahmed2024inferring} contains two main tables (2 and 3) displaying the results; the corresponding replication package contains two folders ({\tt Table2} and {\tt Table3}) containing the data, models, scripts and other information necessary to reproduce the results in the two tables.

Table~\ref{table:rq1} displays the results of running our implementation of DeepInfer on the data and models from the {\tt Table2} folder in our attempt to replicate the results from Table 2 in the DeepInfer paper~\cite{ahmed2024inferring}. Table~\ref{table:rq1} displays the numbers that match in green. 
We summarize our observations below.

\paragraph{DeepInfer} 

We decompose RQ1 into the following two questions:
{\em Are we able to run?} Yes. We were able to run both the original DeepInfer code and our own implementation of it on the datasets provided in the replicated package.
{\em  Did we get the same results? }
Only partially.  For PD and HP, all the numbers match ~\cite{ahmed2024inferring}. 
%
%
%
The results do not match for the BM and GC models. We looked carefully into the reasons for this and also communicated with the authors of DeepInfer. For the BP models, even the ground truth does not match. We also found that the scripts provided in the replicated package removed one of the branches in the code.   In our implementation setup, we run the same code on all the models, avoiding issues such as mentioned above. For GC, we found that the scripts use some pre-computed values that were read from a file. Upon consultation with the authors of DeepInfer, we understand that this was due to variation (randomness) in the inference phase. Although we could not observe such randomness in our experiments, it is  possible that randomness can happen due to slight variations of the setup used in experiments performed by different teams. 

We further attempted to reproduce the results from Table 3 in~\cite{ahmed2024inferring}. However we noticed some issues that we could not resolve. For instance, Table 3 in~\cite{ahmed2024inferring} contains results for ground truth {\bf ActFP} and {\bf ActTP}. ActTP denotes {\em if the actual label and predicted label by a model are not equal} and ActFP denotes {\em if the actual label and predicted label by a model are equal.}
Thus, it appears that ActTP is the same as \# Incorrect while ActFP is the same as \# Correct, yet these numbers seem to be reversed in the Tables in~\cite{ahmed2024inferring}; for instance 
for PD1, Ground Truth \# Correct is 119 in Table 2~\cite{ahmed2024inferring} while Ground Truth ActTP is also 119 in Table 3~\cite{ahmed2024inferring}. Further, the replication package for DeepInfer
contains two folders containing the same pre-trained models (Table2/Models and Table3/Models) with one exception: by computing the file difference, we found that BM3 in Table3/Models is BM11 in Table2/Models and vice-versa. Another issue is that the scripts under Table3 read some results from files for which we could not find the scripts; upon contacting the authors, they explained to us how to reproduce the files by changing some hard coded parameters in the scripts under folder Table2. Given all these issues, we concluded that we do not have sufficient information to be able to reliably reproduce the results in Table 3 ~\cite{ahmed2024inferring}.

\paragraph{SelfChecker} For RQ1, we again aim to answer the following two questions:
{\em Are we able to run?} Yes. We were able to run both the original SelfChecker code and our own implementation of it on the dataset and model provided in the replicated package. 
{\em  Did we get the same results?} Yes we did obtain same results for the TPR, FPR, and F1 metrics with minor differences in the numbers. 
Table~\ref{table:selfchecker} displays our results which reproduce the results from Table 3 in the SelfChecker paper~\cite{Xiao2021SelfCheckingDN}.


\begin{boxC}
\textit{\textbf{Observation 1.} We found it difficult to reproduce the results.
For  DeepInfer we 
found discrepancies due to replication of code across multiple scripts which introduced errors and due to hard-coding of some values (an attempt by the authors to reduce variability in the tool's outputs). 
The unclear definitions for the evaluation metrics used in ~\cite{ahmed2024inferring} also lead to different results.
For SelfChecker we were able to run the tool on the only model that was made available.}
\end{boxC}


\noindent{\bf RQ.2: For the considered approaches, can the results be obtained on artifacts other than their own?}

All three tools are designed to process feed-forward neural networks taking in any type of input. We attempt to run them on both image and tabular data. We use the datasets and models curated using TrustBench for this purpose (details in Table~\ref{tab:tableB}, Table~\ref{tab:tableC} and Table~\ref{table:benchmark}).

\textit{DeepInfer.}
We were not able to run DeepInfer on image data. The tool throws the following error, \texttt{"ValueError: matmul: Input operand 1 has a mismatch in its core dimension 0, with gufunc signature (n?,k),(k,m?)->(n?,m?) (size 3 is different from 32)"}. This indicates that the tool is not able to handle the dimensionality and format of image inputs. 
Consultation with the authors further revealed that the tool, in its current implementation, cannot handle architectures with activation functions in separate individual layers, and the softmax activation function. Therefore, the answer to RQ.2 for DeepInfer is that it cannot be run on image data in its current implementation.

However, we attempted to address this limitation.
We extended DeepInfer to enable application to image datasets,
by building pre-conditions at an inner layer (of lower dimensionality) instead of the input layer; we further modified DeepInfer to skip the layers that it can not handle. This extension enables the tool to run on the CIFAR-10 dataset and model. The results are in Table~\ref{table:rq3_tabular_DI} and will be discussed in RQ.3. 

\textit{SelfChecker.}
We were not able to run SelfChecker on tabular data. The SelfChecker tool throws a 
\texttt{np.linalg.LinAlgError} in the Kernel density estimation function, when run on the models for tabular data. A similar error regarding SelfChecker was reported in ~\cite{ahmed2024inferring} as well.
Therefore, the answer to RQ.2 for SelfChecker is that it cannot be run on tabular data in its current implementation. 

However, we attempted to address this limitation. In our implementation of SelfChecker, 
we expanded the input domain to include tabular data by converting it to the expected format (with Pandas). To resolve the error encountered during KDE generation, we implemented regularization in covariance matrix computation. This regularization ensures the matrix remains non-singular and invertible, with an alpha value of 0.1 determining the regularization strength.  The existing implementation uses Guassian KDE estimation. We specifically modified the kde estimation to use Regularized KDE with the following parameters \verb|bw_method= scott| and \verb|alpha= 0.01|, when the above error was caught. These changes enabled running the tool on the tabular data models without any errors. The results are in Table~\ref{table:rq3_tabular_SC} and will be discussed in RQ.3.

\textit{Prophecy.}
The tool ran without any issues on both image and tabular data, which suffices to answer RQ.2. The results are in Table~\ref{table:rq3_tabular_proph} and will be discussed in RQ.3.

\begin{boxC}
\textit{\textbf{Observation 2.} We found it difficult to produce results for the considered approaches using artifacts other than their own.  DeepInfer and SelfChecker cannot be run on image data and tabular data respectively with their original implementations. We managed to run SelfChecker  on tabular data (and respective models) by changing the KDE function. DeepInfer required a bigger change to enable application to image models. Prophecy was the only approach that worked on both tabular and image data without modification.}
\end{boxC}

\noindent{\bf RQ.3: How do the considered approaches compare in terms
of effectiveness in determining mis-classifications?}

\begin{figure*}[!htbp]
    \centering
    \includegraphics[width=1\textwidth]{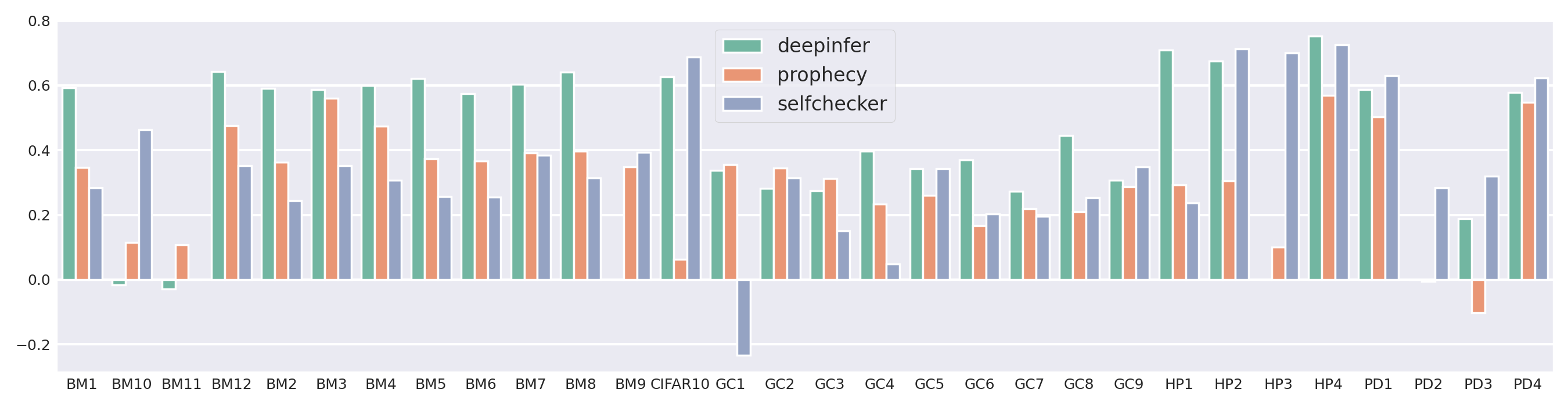}
    \caption{Effectiveness of the approaches in terms of MCC (y-axis) for all models  (x-axis).} 
    \label{fig:effectiveness}
\end{figure*}

We use the same datasets as for RQ2 and the extended versions of DeepInfer and SelfChecker. We use a common definition to calculate the TP, FP, TN, FN metrics for all three approaches as described in section~\ref{subsec:metrics}. Tables ~\ref{table:rq3_tabular_SC}, ~\ref{table:rq3_tabular_DI}, and ~\ref{table:rq3_tabular_proph} presents the results. The appendix has more details. 
We use the MCC metric to compare the effectiveness of the three approaches. Figure~\ref{fig:effectiveness} summarizes the comparison results. 

We present results of only 1 execution, however, we did perform multiple runs for each tool. We found that SelfChecker and DeepInfer produce the same results over multiple runs. Prophecy, on the other hand, displayed variability in the results. We modified the implementation by fixing the "random\_state" parameter in the decision-tree estimator, which made the results stable.



DeepInfer performs well across all the models with tabular data. The average MCC is 0.41 indicating good correlation between the predictions and actuals, except for some cases, specifically for BM10 and BM11 models, where the MCC is slightly negative. It is noteworthy that for HP3, BM9 and PD2, the MCC value is 0 (refer Table~\ref{table:rq3_tabular_DI}. This is to be interpreted as a random correlation, and occurs when both the TP and FP are zeros, or both TN and FN are zeros. In these specific cases, DeepInfer predicts all instances as correct, leading to zero TPs and FPs.

The performance of SelfChecker and Prophecy across all tabular models is comparable, with average MCC of 0.33 and 0.31, respectively. For GC1, SelfChecker performs poorly (MCC of -0.234), where it has zero true positives, and for BM11, SelfChecker has an MCC of 0 since it predicts all instances as incorrect, leading to zero TN and FN (refer Table~\ref{table:rq3_tabular_SC}). Prophecy, on the other hand, does not have zero MCC for any model, indicating that it does not mark all instances as correct or incorrect, for any model. However, it performs poorly on PD2 and PD3 models with negative MCCs (refer Table~\ref{table:rq3_tabular_proph}). 

For the image case-study, both DeepInfer and SelfChecker perform equally well (MCCs of 0.62 and 0.68 respectively). Prophecy performs poorly on the image model with an MCC of 0.06. This could potentially be attributed to the  performance of the decision-tree learning algorithm being adversely impacted by the balance of correctly classified vs mis-classified inputs in the data used in the analyze phase. Given the high accuracy of the ConvNet model on the train data, the efficacy of the mis-classification detection is poor. Enabling Prophecy to use a balanced train dataset in the analysis phase, leads to better performance (MCC 0.497 refer Table~\ref{table:prophecy_balanced}) corroborates our reasoning. 




\begin{boxC}
\textit{\textbf{Observation 3.}  All tools demonstrate meaningful detections with positive MCC values for most models, with DeepInfer performing the best on average. SelfChecker and Prophecy are comparable in their overall effectiveness, with SelfChecker excelling on the image model. DeepInfer and SelfChecker show high variance in effectiveness across models, while Prophecy has a more balanced performance.}
\end{boxC}

\midsepremove
\begin{table}
   \centering
    \begin{scriptsize}
    \caption{Effectiveness results by model for SelfChecker.}
    \label{table:rq3_tabular_SC}
    \begin{tabular}{|c|r r r r | r r r r r|}
    \hline
        \headcol
        & \multicolumn{4}{c|}{\textbf{Confusion Matrix}} & \multicolumn{5}{c|}{\textbf{Metrics}} \\
        \cmidrule{2-5}
        \cmidrule{6-10}
        \headcol
        \multirow{-2}{*}{\textbf{Model}} & \textit{TP} & \textit{FP} & \textit{TN} & \textit{FN} & \textit{TPR} & \textit{FPR} & \textit{Prec.} & \textit{F1} & \textit{MCC} \\ 
        \toprule
        BM1 & 538 & 444 & 76 & 0 & 100\% & 85.38\% & 54.79\% & 70.79\% & 0.283 \\ \rowcol
        BM2 & 536 & 457 & 63 & 2 & 99.63\% & 87.88\% & 53.98\% & 70.02\% & 0.244 \\ 
        BM3 & 534 & 397 & 123 & 4 & 99.26\% & 76.35\% & 57.36\% & 72.7\% & 0.352 \\ \rowcol
        BM4 & 535 & 424 & 96 & 3 & 99.44\% & 81.54\% & 55.79\% & 71.48\% & 0.307 \\ 
        BM5 & 537 & 454 & 66 & 1 & 99.81\% & 87.31\% & 54.19\% & 70.24\% & 0.257 \\ \rowcol
        BM6 & 535 & 450 & 70 & 3 & 99.44\% & 86.54\% & 54.31\% & 70.26\% & 0.255 \\ 
        BM7 & 531 & 372 & 148 & 7 & 98.7\% & 71.54\% & 58.8\% & 73.7\% & 0.384 \\ \rowcol
        BM8 & 535 & 420 & 100 & 3 & 99.44\% & 80.77\% & 56.02\% & 71.67\% & 0.315 \\ 
        BM9 & 534 & 373 & 147 & 4 & 99.26\% & 71.73\% & 58.88\% & 73.91\% & 0.393 \\ \rowcol
        BM10 & 248 & 28 & 492 & 290 & 46.1\% & 5.38\% & 89.86\% & 60.93\% & 0.464 \\ 
        BM11 & 538 & 520 & 0 & 0 & 100.0\% & 100.0\% & 50.85\% & 67.42\% & 0.0 \\ \rowcol
        BM12 & 534 & 397 & 123 & 4 & 99.26\% & 76.35\% & 57.36\% & 72.7\% & 0.352 \\ \midrule
        CIFAR10 & 1244 & 207 & 7838 & 711 & 63.63\% & 2.57\% & 85.73\% & 73.05\% & 0.688 \\ \midrule
        \rowcol
        GC1 & 0 & 10 & 57 & 33 & 0.0\% & 14.93\% & 0.0\% & 0.0\% & -0.234 \\ 
        GC2 & 19 & 17 & 50 & 14 & 57.58\% & 25.37\% & 52.78\% & 55.07\% & 0.315 \\ \rowcol
        GC3 & 22 & 34 & 33 & 11 & 66.67\% & 50.75\% & 39.29\% & 49.44\% & 0.151 \\ 
        GC4 & 17 & 31 & 36 & 16 & 51.52\% & 46.27\% & 35.42\% & 41.98\% & 0.049 \\ \rowcol
        GC5 & 20 & 17 & 50 & 13 & 60.61\% & 25.37\% & 54.05\% & 57.14\% & 0.343 \\
        GC6 & 12 & 12 & 55 & 21 & 36.36\% & 17.91\% & 50.0\% & 42.11\% & 0.203 \\ 
        \rowcol
        GC7 & 13 & 14 & 53 & 20 & 39.39\% & 20.9\% & 48.15\% & 43.33\% & 0.196 \\ 
        GC8 & 13 & 11 & 56 & 20 & 39.39\% & 16.42\% & 54.17\% & 45.61\% & 0.253 \\
        \rowcol
        GC9 & 17 & 12 & 55 & 16 & 51.52\% & 17.91\% & 58.62\% & 54.84\% & 0.348 \\
        \midrule
        HP1 & 72 & 66 & 8 & 0 & 100.0\% & 89.19\% & 52.17\% & 68.57\% & 0.237 \\
        \rowcol
        HP2 & 60 & 9 & 65 & 12 & 83.33\% & 12.16\% & 86.96\% & 85.11\% & 0.713 \\ 
        HP3 & 58 & 8 & 66 & 14 & 80.56\% & 10.81\% & 87.88\% & 84.06\% & 0.701 \\ 
        \rowcol
        HP4 & 62 & 10 & 64 & 10 & 86.11\% & 13.51\% & 86.11\% & 86.11\% & 0.726 \\ 
        \midrule
        PD1 & 26 & 8 & 37 & 6 & 81.25\% & 17.78\% & 76.47\% & 78.79\% & 0.63 \\ 
        \rowcol
        PD2 & 11 & 5 & 40 & 21 & 34.38\% & 11.11\% & 68.75\% & 45.83\% & 0.283 \\
        PD3 & 29 & 28 & 17 & 3 & 90.62\% & 62.22\% & 50.88\% & 65.17\% & 0.319 \\ 
        \rowcol
        PD4 & 24 & 6 & 39 & 8 & 75.0\% & 13.33\% & 80.0\% & 77.42\% & 0.623 \\ 
        \bottomrule
    \end{tabular}
     \end{scriptsize}
\end{table}

\begin{table}
    \begin{scriptsize}
    \caption{Effectiveness results by model for DeepInfer.}
    \label{table:rq3_tabular_DI}
  
    \begin{tabular}{|c| r r r r | r r r r r|}
    \hline
        \headcol
        & \multicolumn{4}{c|}{\textbf{Confusion Matrix}} & \multicolumn{5}{c|}{\textbf{Metrics}} \\
        \cmidrule{2-5}
        \cmidrule{6-10}
        \headcol
        \multirow{-2}{*}{\textbf{Model}}& \textit{TP} & \textit{FP} & \textit{TN} & \textit{FN} & \textit{TPR} & \textit{FPR} & \textit{Prec.} & \textit{F1} & \textit{MCC} \\ 
        \toprule
        BM1& 597 & 151 & 259 & 51 & 92.13\% & 36.83\% & 79.81\% & 85.53\% & 0.592 \\ \rowcol
        BM2& 246 & 59 & 617 & 136 & 64.4\% & 8.73\% & 80.66\% & 71.62\% & 0.59 \\
        BM3& 560 & 138 & 290 & 70 & 88.89\% & 32.24\% & 80.23\% & 84.34\% & 0.587 \\ \rowcol
        BM4& 422 & 93 & 424 & 119 & 78.0\% & 17.99\% & 81.94\% & 79.92\% & 0.6 \\
        BM5& 427 & 125 & 429 & 77 & 84.72\% & 22.56\% & 77.36\% & 80.87\% & 0.621 \\ \rowcol
        BM6& 611 & 146 & 243 & 58 & 91.33\% & 37.53\% & 80.71\% & 85.69\% & 0.575 \\ 
        BM7& 288 & 71 & 573 & 126 & 69.57\% & 11.02\% & 80.22\% & 74.51\% & 0.603 \\ \rowcol
        BM8& 428 & 90 & 440 & 100 & 81.06\% & 16.98\% & 82.63\% & 81.84\% & 0.641 \\ 
        BM9& 0 & 0 & 857 & 201 & 0.0\% & 0.0\% & 0.0\% & 0.0\% & 0.0 \\
        \rowcol
        BM10& 260 & 249 & 260 & 289 & 47.36\% & 48.92\% & 51.08\% & 49.15\% & -0.016 \\ 
        BM11& 321 & 292 & 199 & 246 & 56.61\% & 59.47\% & 52.37\% & 54.41\% & -0.029 \\ \rowcol
        BM12& 424 & 99 & 445 & 90 & 82.49\% & 18.2\% & 81.07\% & 81.77\% & 0.643 \\ \midrule
        CIFAR10& 3938 & 1492 & 4107 & 463 & 89.48\% & 26.65\% & 72.52\% & 80.11\% & 0.626 \\
        \midrule
        \rowcol
        GC1& 30 & 18 & 37 & 15 & 66.67\% & 32.73\% & 62.5\% & 64.52\% & 0.338 \\
        GC2& 53 & 26 & 14 & 7 & 88.33\% & 65.0\% & 67.09\% & 76.26\% & 0.281 \\
        \rowcol
        GC3& 16 & 10 & 51 & 23 & 41.03\% & 16.39\% & 61.54\% & 49.23\% & 0.274 \\ 
        GC4& 38 & 6 & 29 & 27 & 58.46\% & 17.14\% & 86.36\% & 69.72\% & 0.397 \\ 
        \rowcol
        GC5& 23 & 9 & 44 & 24 & 48.94\% & 16.98\% & 71.88\% & 58.23\% & 0.342 \\
        GC6& 40 & 8 & 27 & 25 & 61.54\% & 22.86\% & 83.33\% & 70.8\% & 0.369 \\
        \rowcol
        GC7& 50 & 21 & 17 & 12 & 80.65\% & 55.26\% & 70.42\% & 75.19\% & 0.272 \\ 
        GC8& 33 & 3 & 34 & 30 & 52.38\% & 8.11\% & 91.67\% & 66.67\% & 0.445 \\
        \rowcol
        GC9& 45 & 14 & 22 & 19 & 70.31\% & 38.89\% & 76.27\% & 73.17\% & 0.307 \\ \midrule
        HP1& 64 & 5 & 60 & 17 & 79.01\% & 7.69\% & 92.75\% & 85.33\% & 0.71 \\
        \rowcol
        HP2& 56 & 8 & 66 & 16 & 77.78\% & 10.81\% & 87.5\% & 82.35\% & 0.675 \\
        HP3& 0 & 0 & 74 & 72 & 0.0\% & 0.0\% & 0.0\% & 0.0\% & 0.0 \\ 
        \rowcol
        HP4& 60 & 8 & 68 & 10 & 85.71\% & 10.53\% & 88.24\% & 86.96\% & 0.753 \\ 
        \midrule
        PD1& 21 & 3 & 40 & 13 & 61.76\% & 6.98\% & 87.5\% & 72.41\% & 0.587 \\
        \rowcol
        PD2& 0 & 0 & 45 & 32 & 0.0\% & 0.0\% & 0.0\% & 0.0\% & 0.0 \\ 
        PD3& 19 & 17 & 27 & 14 & 57.58\% & 38.64\% & 52.78\% & 55.07\% & 0.188 \\ \rowcol
        PD4& 45 & 12 & 17 & 3 & 93.75\% & 41.38\% & 78.95\% & 85.71\% & 0.579 \\
        \bottomrule
    \end{tabular}
    \end{scriptsize}
\end{table}

\begin{table}
   \centering
    \begin{scriptsize}
    \caption{Effectiveness results by model for Prophecy.}
    \label{table:rq3_tabular_proph}
    \begin{tabular}{|c|r|r|r|r|r|r|r|r|r|r|r|}
    \hline
        \headcol
        & \multicolumn{4}{c|}{\textbf{Confusion Matrix}} & \multicolumn{5}{c|}{\textbf{Metrics}} \\
        \cmidrule{2-5}
        \cmidrule{6-10}
        \headcol
        \multirow{-2}{*}{\textbf{Model}} & \textit{TP} & \textit{FP} & \textit{TN} & \textit{FN} & \textit{TPR} & \textit{FPR} & \textit{Prec.} & \textit{F1} & \textit{MCC} \\ 
        \toprule
        BM1 & 72 & 42 & 784 & 160 & 31.03\% & 5.08\% & 63.16\% & 41.62\% & 0.346 \\
        \rowcol
        BM2 & 75 & 43 & 788 & 152 & 33.04\% & 5.17\% & 63.56\% & 43.48\% & 0.363 \\
        BM3 & 253 & 98 & 597 & 110 & 69.7\% & 14.1\% & 72.08\% & 70.87\% & 0.561 \\
        \rowcol
        BM4 & 151 & 60 & 695 & 152 & 49.83\% & 7.95\% & 71.56\% & 58.75\% & 0.474 \\ 
        BM5 & 81 & 41 & 775 & 161 & 33.47\% & 5.02\% & 66.39\% & 44.51\% & 0.374 \\
        \rowcol
        BM6 & 85 & 52 & 769 & 152 & 35.86\% & 6.33\% & 62.04\% & 45.45\% & 0.367 \\ 
        BM7 & 88 & 47 & 773 & 150 & 36.97\% & 5.73\% & 65.19\% & 47.18\% & 0.391 \\ 
        \rowcol
        BM8& 94 & 67 & 774 & 123 & 43.32\% & 7.97\% & 58.39\% & 49.74\% & 0.397 \\
        BM9& 78 & 55 & 779 & 146 & 34.82\% & 6.59\% & 58.65\% & 43.7\% & 0.348 \\ 
        \rowcol
        BM10& 221 & 454 & 299 & 84 & 72.46\% & 60.29\% & 32.74\% & 45.1\% & 0.115 \\ 
        BM11& 229 & 449 & 291 & 89 & 72.01\% & 60.68\% & 33.78\% & 45.98\% & 0.108 \\
        \rowcol
        BM12& 130 & 65 & 739 & 124 & 51.18\% & 8.08\% & 66.67\% & 57.91\% & 0.475 \\ 
        \midrule
        CIFAR10& 91 & 186 & 7954 & 1769 & 4.89\% & 2.29\% & 32.85\% & 8.52\% & 0.062 \\ 
        \midrule
        \rowcol
        GC1& 35 & 22 & 32 & 11 & 76.09\% & 40.74\% & 61.4\% & 67.96\% & 0.356 \\ 
        GC2& 27 & 22 & 40 & 11 & 71.05\% & 35.48\% & 55.1\% & 62.07\% & 0.345 \\ 
        \rowcol
        GC3& 23 & 14 & 44 & 19 & 54.76\% & 24.14\% & 62.16\% & 58.23\% & 0.313 \\ 
        GC4& 14 & 12 & 53 & 21 & 40.0\% & 18.46\% & 53.85\% & 45.9\% & 0.234 \\ \rowcol
        GC5& 17 & 18 & 50 & 15 & 53.12\% & 26.47\% & 48.57\% & 50.75\% & 0.261 \\
        GC6& 10 & 12 & 57 & 21 & 32.26\% & 17.39\% & 45.45\% & 37.74\% & 0.166 \\ 
        GC7& 13 & 12 & 54 & 21 & 38.24\% & 18.18\% & 52.0\% & 44.07\% & 0.219 \\ 
        \rowcol
        GC8& 9 & 6 & 58 & 27 & 25.0\% & 9.38\% & 60.0\% & 35.29\% & 0.21 \\ 
        GC9& 20 & 16 & 47 & 17 & 54.05\% & 25.4\% & 55.56\% & 54.79\% & 0.288 \\ \midrule
        HP1& 6 & 6 & 118 & 16 & 27.27\% & 4.84\% & 50.0\% & 35.29\% & 0.292 \\ 
        \rowcol
        HP2& 7 & 6 & 115 & 18 & 28.0\% & 4.96\% & 53.85\% & 36.84\% & 0.305 \\ 
        HP3& 19 & 62 & 55 & 10 & 65.52\% & 52.99\% & 23.46\% & 34.55\% & 0.101 \\ \rowcol
        HP4& 16 & 7 & 112 & 11 & 59.26\% & 5.88\% & 69.57\% & 64.0\% & 0.569 \\ 
        \midrule
        PD& 11 & 2 & 50 & 14 & 44.0\% & 3.85\% & 84.62\% & 57.89\% & 0.502 \\
        \rowcol
        PD2& 5 & 9 & 40 & 23 & 17.86\% & 18.37\% & 35.71\% & 23.81\% & -0.006 \\ 
        PD3& 2 & 8 & 44 & 23 & 8.0\% & 15.38\% & 20.0\% & 11.43\% & -0.103 \\
        \rowcol
        PD4& 15 & 4 & 47 & 11 & 57.69\% & 7.84\% & 78.95\% & 66.67\% & 0.547 \\ \bottomrule
    \end{tabular}
     \end{scriptsize}
\end{table}

\begin{table}
    \centering
    \begin{scriptsize}
    \caption{Effectiveness of Prophecy on CIFAR10 with balanced train data.}
    \label{table:prophecy_balanced}
    \begin{tabular}{|c | r r r r | r r r r r |}
    \toprule
    \headcol
    & \multicolumn{4}{c|}{\textbf{Confusion Matrix}} & \multicolumn{5}{c|}{\textbf{Metrics}} \\
    \cmidrule{2-5}
    \cmidrule{6-10}
    \headcol
    \multirow{-2}{*}{\textbf{Model}} & \textit{TP} & \textit{FP} & \textit{TN} & \textit{FN} & \textit{TPR} & \textit{FPR} & \textit{Precision} &\textit{F1} & \textit{MCC} \\
    \midrule
    \cellcolor{lavender} \textit{CIFAR10} & 1511 & 1423 & 6534 & 532 & 73.96\% & 17.88\% & 51.5\% & 60.72\% & 0.497 \\
    \rowcol
    \bottomrule
    \end{tabular}
    \end{scriptsize}
\end{table}
\midsepdefault
\midsepremove
\begin{table}
{\scriptsize
    \centering
    \caption{Time (average duration in seconds) and Memory (peak memory usage in Mebibytes) efficiency for each approach by dataset.}
    \label{tab:efficiency}
    \begin{tabular}{| c | c | c | r | r |}
    \hline
        \headcol
        \textbf{Tool} & \textbf{Phase} & \textbf{Dataset} & \textbf{Duration} & \textbf{Memory} \\ 
        \toprule
        \cellcolor{lavender} & analyze & BM & 3.45 & 687.58 \\
        \rowcol
         & analyze & CIFAR10 & 55.72 & 29 150.96 \\
        \cellcolor{lavender} & analyze & GC & 4.89 & 661.32 \\
        \rowcol
        & analyze & HP & 2.82 & 643.71 \\
        \cellcolor{lavender} & analyze & PD & 2.67 & 659.31 \\
        \rowcol
        & infer & BM & 2.63 & 646.03 \\
        \cellcolor{lavender} & infer & CIFAR10 & 29.01 &
        29 146.96 \\
        \rowcol
        & infer & GC & 2.64 & 642.5 \\
        \cellcolor{lavender} & infer & HP & 2.62 & 646.69 \\
        \rowcol
        \multirow{-10}{*}{\textit{DeepInfer}} & infer & PD & 2.67 & 648.37 \\
        \midrule
        \cellcolor{lavender} & analyze & BM & 3.94 & 730.33 \\
        \rowcol
        & analyze & CIFAR10 & 118.42 & 4 177.04 \\
        \cellcolor{lavender}& analyze & GC & 3.33 & 711.58 \\
        \rowcol
        & analyze & HP & 3.22 & 709.97 \\
        \cellcolor{lavender}& analyze & PD & 3.27 & 710.97 \\
        \rowcol
        & infer & BM & 3.49 & 696.4 \\
        \cellcolor{lavender}& infer & CIFAR10 & 25.97 & 2 012.66 \\
        \rowcol
        & infer & GC & 3.04 & 696.16 \\
        \cellcolor{lavender} & infer & HP & 3.07 & 699.33 \\
        \rowcol
        \multirow{-10}{*}{\textit{Prophecy}} & infer & PD & 3.02 & 699.5 \\
        \midrule
        \cellcolor{lavender} & analyze & BM & 9.05 & 4 063.45 \\
        \rowcol
         & analyze & CIFAR10 & 1 930.52 & 40 871.41 \\
        \cellcolor{lavender} & analyze & GC & 3.51 & 3 959.27 \\
         \rowcol
         & analyze & HP & 3.47 & 4 069.1 \\
         \cellcolor{lavender} & analyze & PD & 3.52 & 3 902.47 \\
         \rowcol
         & infer & BM & 4.44 & 3 773.3 \\
         \cellcolor{lavender} & infer & CIFAR10 & 1 352.59 & 7 363.5 \\
         \rowcol
         & infer & GC & 3.22 & 3 796.63 \\
         \cellcolor{lavender} & infer & HP & 3.17 & 3 833.67 \\
        \rowcol
        \multirow{-10}{*}{\textit{SelfChecker}} & infer & PD & 3.17 & 3 758.48 \\
        \bottomrule
    \end{tabular}
    }
\end{table}

\noindent{\bf RQ.4: How do the considered approaches compare in terms
of time and memory consumption?}

Our evaluation of the efficiency of three approaches—DeepInfer, Prophecy, and SelfChecker—revealed negligible differences in total execution duration (both \textit{analyze} and \textit{infer} phases) on the models with tabular data (average 3.5 secs). These models are relatively small in size (refer table~\ref{table:benchmark}). However, the times were longer on the bigger ConvNet model for CIFAR10, with SelfChecker consuming the maximum amount of time (~32.2 minutes).  
In terms of the memory consumption, SelfChecker again seems to be very expensive for both types of models (average peak memory usage of 4K Mebibytes for tabular models and 41K Mebibytes for the image model). The memory usage of DeepInfer is lower for tabular models but jumps to 29K Mebibytes for the image model. Prophecy has overall low memory usage for all models.
Table~5 in the appendix (see full paper in replication package) presents the details.

\begin{boxC}
\textit{\textbf{Observation 4.} Overall, we can observe that SelfChecker seems to be the most resource intensive, while Prophecy seems to be the least across all models. }
\end{boxC}




\section{Discussion}

\subsection{Threats to Validity}

A tool tasked with assessing the reliability of a pre-trained model needs to execute the model on data "unseen" by the model. This requires knowledge of the exact train/val/test data used to train the model. We are uncertain of the overlap between the train set used to train the model and the data used to train the detector. This can add a threat to the validity of our results. However, we envisage this to be typically the case when handling off-the-shelf models.

Our code modifications, which consist of refactorings, optimizations and extensions, may have introduced errors; we addressed by being able to reproduce the results for DeepInfer (partially) and SelfChecker (fully). 
Although we performed preliminary experiments to analyze variability of results over multiple executions, this may not be sufficient to make reliable conclusions. 
Further, the modification made to Prophecy's code to stabilize its results introduces threat to validity. 
The number and type of datasets and models used threatens the external validity of our results regarding the tools.

\subsection{Lessons Learned}

Our study emphasizes again the need for open science. 
Beyond the application considered in this study, in order to replicate and evaluate tools in general, we would like to emphasize on the following key characteristics.
\begin{itemize}
\item \textit{Transparency:} Tool repositories should store precise artifacts that act as inputs (datasets, models so on), store experimental results, add proper documentation explaining usage and precise configuration parameters.
\item \textit{Generalizability and Extensibility:} In order to apply the tool to various benchmarks, the code should avoid hard-coding parameters. The code should be modular and not tightly coupled.
The repository structure should also be modular, avoid duplication of scripts, and apply same code consistently across all benchmarks.
\item \textit{Handling variability in results:} Although we did not notice variability in our study, the inherent randomness of machine learning models necessitates a large number of trials to ensure reliability of the results. We also recommend using a docker technology when building replication package.
\end{itemize}

We hope the \textit{TrustBench} and the \textit{TrustDNN} framework aid in addressing some of the lessons learned for future tools.




\section{Conclusion}

In this paper, we evaluated recent approaches that have been proposed for evaluating the reliability of DNNs. We found that it is difficult to run and reproduce the
results for these approaches on their replication packages, and it
is also difficult to run them on artifacts other than their
own. Further, it is difficult to compare the effectiveness of the approaches,
as they use different evaluation metrics. Our results indicate that more effort is needed in our research community to obtain
sound techniques for evaluating the reliability of DNNs. To this end, we contribute an evaluation framework that we make available as open source. The framework 
enables experimentation with different approaches for evaluating reliability of DNNs, allowing comparison using clear metrics. We hope that the community can build on the framework that we provide and continue research on this challenging topic.

\begin{acks}
The first author was supported by CMU Portugal through
national funds under the grant PRT/BD/152197/2021~\cite{FCT2021}. We thank Ahmed Shibbir for the insightful feedback he provided on the replication package for DeepInfer.
\end{acks}

\newpage
\bibliographystyle{ACM-Reference-Format}
\bibliography{refs}
\pagebreak

\section*{Appendix}

\lstset{style=jsonstyle}
\vspace{0.3cm}

\begin{minipage}{0.5\textwidth}
\centering
\begin{lstlisting}[language=json, caption={Configuration of each tool in TrustDNN, for the replication experiments.},label={lst:replication},frame=tlrb, basicstyle=\fontfamily{pcr}\selectfont\small]
        DeepInfer:{
            "condition": ">=",
            "prediction_interval": 0.95
        }
        SelfChecker:{
            "var_threshold": 1e-5,
            "only_activation_layers": true,
            "batch_size": 128
        }
\end{lstlisting}

\begin{lstlisting}[language=json, caption={Configuration of each tool in TrustDNN, for comparison experiments (using both only\_dense\_layers and only\_activation\_layers flags selects dense layers coupled with activation functions.)},label={lst:comparison},frame=tlrb, basicstyle=\fontfamily{pcr}\selectfont\small]
        Prophecy: {
            "only_activation_layers": true,
            "only_dense_layers": true,
            "random_state": 42,
            "skip_rules": true
        }
        DeepInfer:{
            "condition": ">=",
            "prediction_interval": 0.95
        }
        SelfChecker:{
            "var_threshold": 1e-5,
            "only_activation_layers": true,
            "only_dense_layers": true,
            "batch_size": 128
        }
\end{lstlisting}
\end{minipage}

\begin{table*}[htbp]
    \centering
    \caption{Effectiveness results by model for Prophecy.}
    \label{table:rq3_tabular_proph}
    \begin{tabular}{|l|r|r|r|r|r|r|r|r|r|r|r|r|r|r|}
    \hline
        \headcol
        & \multicolumn{3}{c|}{\textbf{Notifications}} & \multicolumn{4}{c|}{\textbf{Confusion Matrix}} & \multicolumn{5}{c|}{\textbf{Metrics}} \\
        \cmidrule{2-4}
        \cmidrule{5-8}
        \cmidrule{9-13}
        \headcol
        \multirow{-2}{*}{\textbf{Model}} & \textit{\#C} & \textit{\#I} & \textit{\#U} & \textit{TP} & \textit{FP} & \textit{TN} & \textit{FN} & \textit{TPR} & \textit{FPR} & \textit{Prec.} & \textit{F1} & \textit{MCC} \\ 
        \toprule
        BM1 & 944& 114& 0& 72 & 42 & 784 & 160 & 31.03\% & 5.08\% & 63.16\% & 41.62\% & 0.346 \\
        \rowcol
        BM2 & 940& 118& 0& 75 & 43 & 788 & 152 & 33.04\% & 5.17\% & 63.56\% & 43.48\% & 0.363 \\
        BM3 & 707& 60& 291& 253 & 98 & 597 & 110 & 69.7\% & 14.1\% & 72.08\% & 70.87\% & 0.561 \\
        \rowcol
        BM4 & 847& 92& 119& 151 & 60 & 695 & 152 & 49.83\% & 7.95\% & 71.56\% & 58.75\% & 0.474 \\ 
        BM5 & 936& 122& 0& 81 & 41 & 775 & 161 & 33.47\% & 5.02\% & 66.39\% & 44.51\% & 0.374 \\
        \rowcol
        BM6 & 921& 137& 0& 85 & 52 & 769 & 152 & 35.86\% & 6.33\% & 62.04\% & 45.45\% & 0.367 \\ 
        BM7 & 923& 135& 0& 88 & 47 & 773 & 150 & 36.97\% & 5.73\% & 65.19\% & 47.18\% & 0.391 \\ 
        \rowcol
        BM8 & 897& 93& 68& 94 & 67 & 774 & 123 & 43.32\% & 7.97\% & 58.39\% & 49.74\% & 0.397 \\
        BM9 & 925& 133& 0& 78 & 55 & 779 & 146 & 34.82\% & 6.59\% & 58.65\% & 43.7\% & 0.348 \\ 
        \rowcol
        BM10 & 383& 675& 0& 221 & 454 & 299 & 84 & 72.46\% & 60.29\% & 32.74\% & 45.1\% & 0.115 \\ 
        BM11 & 380& 678& 0& 229 & 449 & 291 & 89 & 72.01\% & 60.68\% & 33.78\% & 45.98\% & 0.108 \\
        \rowcol
        BM12 & 863& 82& 113& 130 & 65 & 739 & 124 & 51.18\% & 8.08\% & 66.67\% & 57.91\% & 0.475 \\ 
        \midrule
        CIFAR10 & 9723& 101& 176& 91 & 186 & 7954 & 1769 & 4.89\% & 2.29\% & 32.85\% & 8.52\% & 0.062 \\ 
        \midrule
        \rowcol
        GC1 & 43& 9& 48& 35 & 22 & 32 & 11 & 76.09\% & 40.74\% & 61.4\% & 67.96\% & 0.356 \\ 
        GC2 & 51& 11& 38& 27 & 22 & 40 & 11 & 71.05\% & 35.48\% & 55.1\% & 62.07\% & 0.345 \\ 
        \rowcol
        GC3 & 63& 9& 28& 23 & 14 & 44 & 19 & 54.76\% & 24.14\% & 62.16\% & 58.23\% & 0.313 \\ 
        GC4 & 74& 26& 0& 14 & 12 & 53 & 21 & 40.0\% & 18.46\% & 53.85\% & 45.9\% & 0.234 \\ \rowcol
        GC5 & 65& 23& 12& 17 & 18 & 50 & 15 & 53.12\% & 26.47\% & 48.57\% & 50.75\% & 0.261 \\
        GC6 & 78& 22& 0& 10 & 12 & 57 & 21 & 32.26\% & 17.39\% & 45.45\% & 37.74\% & 0.166 \\ 
        GC7 & 75& 25& 0& 13 & 12 & 54 & 21 & 38.24\% & 18.18\% & 52.0\% & 44.07\% & 0.219 \\ 
        \rowcol
        GC8 & 85& 15& 0& 9 & 6 & 58 & 27 & 25.0\% & 9.38\% & 60.0\% & 35.29\% & 0.21 \\ 
        GC9 & 64& 11& 25& 20 & 16 & 47 & 17 & 54.05\% & 25.4\% & 55.56\% & 54.79\% & 0.288 \\ \midrule
        HP1 & 134& 12& 0& 6 & 6 & 118 & 16 & 27.27\% & 4.84\% & 50.0\% & 35.29\% & 0.292 \\ 
        \rowcol
        HP2 & 133& 13& 0& 7 & 6 & 115 & 18 & 28.0\% & 4.96\% & 53.85\% & 36.84\% & 0.305 \\ 
        HP3 & 65& 81& 0& 19 & 62 & 55 & 10 & 65.52\% & 52.99\% & 23.46\% & 34.55\% & 0.101 \\ \rowcol
        HP4 & 123& 8& 15& 16 & 7 & 112 & 11 & 59.26\% & 5.88\% & 69.57\% & 64.0\% & 0.569 \\ 
        \midrule
        PD1 & 64& 13& 0& 11 & 2 & 50 & 14 & 44.0\% & 3.85\% & 84.62\% & 57.89\% & 0.502 \\
        \rowcol
        PD2 & 63& 14& 0& 5 & 9 & 40 & 23 & 17.86\% & 18.37\% & 35.71\% & 23.81\% & -0.006 \\ 
        PD3 & 67& 10& 0& 2 & 8 & 44 & 23 & 8.0\% & 15.38\% & 20.0\% & 11.43\% & -0.103 \\
        \rowcol
        PD4 & 58 & 9& 10& 15 & 4 & 47 & 11 & 57.69\% & 7.84\% & 78.95\% & 66.67\% & 0.547 \\ \bottomrule
    \end{tabular}
\end{table*}

\begin{table*}[htbp]
    \centering
    \caption{Effectiveness of Prophecy on CIFAR10 with balanced train data.}
    \label{table:prophecy_balanced}
    \begin{tabular}{c | r r r r | r r r r r }
    \toprule
    \headcol
    & \multicolumn{4}{c|}{\textbf{Confusion Matrix}} & \multicolumn{5}{c}{\textbf{Metrics}} \\
    \cmidrule{2-5}
    \cmidrule{6-10}
    \headcol
    \multirow{-2}{*}{\textbf{Model}} & \textit{TP} & \textit{FP} & \textit{TN} & \textit{FN} & \textit{TPR} & \textit{FPR} & \textit{Precision} &\textit{F1} & \textit{MCC} \\
    \midrule
    \cellcolor{lavender} \textit{CIFAR10} & 1511 & 1423 & 6534 & 532 & 73.96\% & 17.88\% & 51.5\% & 60.72\% & 0.497 \\
    \rowcol
    \bottomrule
    \end{tabular}
\end{table*}


\begin{table*}[htbp]
    \centering
    \caption{Effectiveness results by model for DeepInfer.}
    \label{table:rq3_tabular_DI}
    \begin{tabular}{|c| r r r | r r r r | r r r r r|}
    \hline
        \headcol
        & \multicolumn{3}{c|}{\textbf{Notifications}} & \multicolumn{4}{c|}{\textbf{Confusion Matrix}} & \multicolumn{5}{c|}{\textbf{Metrics}} \\
        \cmidrule{2-4}
        \cmidrule{5-8}
        \cmidrule{9-13}
        \headcol
        \multirow{-2}{*}{\textbf{Model}} & \textit{\#C} & \textit{\#I} & \textit{\#U} & \textit{TP} & \textit{FP} & \textit{TN} & \textit{FN} & \textit{TPR} & \textit{FPR} & \textit{Prec.} & \textit{F1} & \textit{MCC} \\ 
        \toprule
        BM1 & 310 & 748 & 0 & 597 & 151 & 259 & 51 & 92.13\% & 36.83\% & 79.81\% & 85.53\% & 0.592 \\ \rowcol
        BM2 & 753 & 305 & 0 & 246 & 59 & 617 & 136 & 64.4\% & 8.73\% & 80.66\% & 71.62\% & 0.59 \\
        BM3 & 360 & 698 & 0 & 560 & 138 & 290 & 70 & 88.89\% & 32.24\% & 80.23\% & 84.34\% & 0.587 \\ \rowcol
        BM4 & 543 & 515 & 0 & 422 & 93 & 424 & 119 & 78.0\% & 17.99\% & 81.94\% & 79.92\% & 0.6 \\
        BM5 & 506 & 552 & 0 & 427 & 125 & 429 & 77 & 84.72\% & 22.56\% & 77.36\% & 80.87\% & 0.621 \\ \rowcol
        BM6 & 301 & 757 & 0 & 611 & 146 & 243 & 58 & 91.33\% & 37.53\% & 80.71\% & 85.69\% & 0.575 \\ 
        BM7 & 699 & 359 & 0 & 288 & 71 & 573 & 126 & 69.57\% & 11.02\% & 80.22\% & 74.51\% & 0.603 \\ \rowcol
        BM8 & 540 & 518 & 0 & 428 & 90 & 440 & 100 & 81.06\% & 16.98\% & 82.63\% & 81.84\% & 0.641 \\ 
        BM9 & 1058 & 0 & 0 & 0 & 0 & 857 & 201 & 0.0\% & 0.0\% & 0.0\% & 0.0\% & 0.0 \\
        \rowcol
        BM10 & 549 & 509 & 0 & 260 & 249 & 260 & 289 & 47.36\% & 48.92\% & 51.08\% & 49.15\% & -0.016 \\ 
        BM11 & 445 & 613 & 0 & 321 & 292 & 199 & 246 & 56.61\% & 59.47\% & 52.37\% & 54.41\% & -0.029 \\ \rowcol
        BM12 & 535 & 523 & 0 & 424 & 99 & 445 & 90 & 82.49\% & 18.2\% & 81.07\% & 81.77\% & 0.643 \\ \midrule
        CIFAR10 & 4570 & 5430 & 0 & 3938 & 1492 & 4107 & 463 & 89.48\% & 26.65\% & 72.52\% & 80.11\% & 0.626 \\
        \midrule
        \rowcol
        GC1 & 52 & 48 & 0 & 30 & 18 & 37 & 15 & 66.67\% & 32.73\% & 62.5\% & 64.52\% & 0.338 \\
        GC2 & 21 & 79 & 0 & 53 & 26 & 14 & 7 & 88.33\% & 65.0\% & 67.09\% & 76.26\% & 0.281 \\
        \rowcol
        GC3 & 74 & 26 & 0 & 16 & 10 & 51 & 23 & 41.03\% & 16.39\% & 61.54\% & 49.23\% & 0.274 \\ 
        GC4 & 56 & 44 & 0 & 38 & 6 & 29 & 27 & 58.46\% & 17.14\% & 86.36\% & 69.72\% & 0.397 \\ 
        \rowcol
        GC5 & 68 & 32 & 0 & 23 & 9 & 44 & 24 & 48.94\% & 16.98\% & 71.88\% & 58.23\% & 0.342 \\
        GC6 & 52 & 48 & 0 & 40 & 8 & 27 & 25 & 61.54\% & 22.86\% & 83.33\% & 70.8\% & 0.369 \\
        \rowcol
        GC7 & 29 & 71 & 0 & 50 & 21 & 17 & 12 & 80.65\% & 55.26\% & 70.42\% & 75.19\% & 0.272 \\ 
        GC8 & 64 & 36 & 0 & 33 & 3 & 34 & 30 & 52.38\% & 8.11\% & 91.67\% & 66.67\% & 0.445 \\
        \rowcol
        GC9 & 41 & 59 & 0 & 45 & 14 & 22 & 19 & 70.31\% & 38.89\% & 76.27\% & 73.17\% & 0.307 \\ \midrule
        HP1 & 77 & 68 & 1 & 64 & 5 & 60 & 17 & 79.01\% & 7.69\% & 92.75\% & 85.33\% & 0.71 \\
        \rowcol
        HP2 & 82 & 64 & 0 & 56 & 8 & 66 & 16 & 77.78\% & 10.81\% & 87.5\% & 82.35\% & 0.675 \\
        HP3 & 146 & 0 & 0 & 0 & 0 & 74 & 72 & 0.0\% & 0.0\% & 0.0\% & 0.0\% & 0.0 \\ 
        \rowcol
        HP4 & 78 & 68 & 0 & 60 & 8 & 68 & 10 & 85.71\% & 10.53\% & 88.24\% & 86.96\% & 0.753 \\ 
        \midrule
        PD1 & 53 & 22 & 2 & 21 & 3 & 40 & 13 & 61.76\% & 6.98\% & 87.5\% & 72.41\% & 0.587 \\
        \rowcol
        PD2 & 77 & 0 & 0 & 0 & 0 & 45 & 32 & 0.0\% & 0.0\% & 0.0\% & 0.0\% & 0.0 \\ 
        PD3 & 41 & 36 & 0 & 19 & 17 & 27 & 14 & 57.58\% & 38.64\% & 52.78\% & 55.07\% & 0.188 \\ \rowcol
        PD4 & 20 & 57 & 0 & 45 & 12 & 17 & 3 & 93.75\% & 41.38\% & 78.95\% & 85.71\% & 0.579 \\
        \bottomrule
    \end{tabular}
\end{table*}

\begin{table*}[htbp]
    \centering
    \caption{Effectiveness results by model for SelfChecker.}
    \label{table:rq3_tabular_SC}
    \begin{tabular}{|c| r r r | r r r r | r r r r r|}
    \hline
        \headcol
        & \multicolumn{3}{c|}{\textbf{Notifications}} & \multicolumn{4}{c|}{\textbf{Confusion Matrix}} & \multicolumn{5}{c|}{\textbf{Metrics}} \\
        \cmidrule{2-4}
        \cmidrule{5-8}
        \cmidrule{9-13}
        \headcol
        \multirow{-2}{*}{\textbf{Model}} & \textit{\#C} & \textit{\#I} & \textit{\#U} & \textit{TP} & \textit{FP} & \textit{TN} & \textit{FN} & \textit{TPR} & \textit{FPR} & \textit{Prec.} & \textit{F1} & \textit{MCC} \\ 
        \toprule
        BM1 & - & - & - & 538 & 444 & 76 & 0 & 100\% & 85.38\% & 54.79\% & 70.79\% & 0.283 \\ \rowcol
        BM2 & - & - & - & 536 & 457 & 63 & 2 & 99.63\% & 87.88\% & 53.98\% & 70.02\% & 0.244 \\ 
        BM3 & - & - & - & 534 & 397 & 123 & 4 & 99.26\% & 76.35\% & 57.36\% & 72.7\% & 0.352 \\ \rowcol
        BM4 & - & - & - & 535 & 424 & 96 & 3 & 99.44\% & 81.54\% & 55.79\% & 71.48\% & 0.307 \\ 
        BM5 & - & - & - & 537 & 454 & 66 & 1 & 99.81\% & 87.31\% & 54.19\% & 70.24\% & 0.257 \\ \rowcol
        BM6 & - & - & - & 535 & 450 & 70 & 3 & 99.44\% & 86.54\% & 54.31\% & 70.26\% & 0.255 \\ 
        BM7 & - & - & - & 531 & 372 & 148 & 7 & 98.7\% & 71.54\% & 58.8\% & 73.7\% & 0.384 \\ \rowcol
        BM8 & - & - & - & 535 & 420 & 100 & 3 & 99.44\% & 80.77\% & 56.02\% & 71.67\% & 0.315 \\ 
        BM9 & - & - & - & 534 & 373 & 147 & 4 & 99.26\% & 71.73\% & 58.88\% & 73.91\% & 0.393 \\ \rowcol
        BM10 & - & - & - & 248 & 28 & 492 & 290 & 46.1\% & 5.38\% & 89.86\% & 60.93\% & 0.464 \\ 
        BM11 & - & - & - & 538 & 520 & 0 & 0 & 100.0\% & 100.0\% & 50.85\% & 67.42\% & 0.0 \\ \rowcol
        BM12 & - & - & - & 534 & 397 & 123 & 4 & 99.26\% & 76.35\% & 57.36\% & 72.7\% & 0.352 \\ \midrule
        CIFAR10 & - & - & - & 1244 & 207 & 7838 & 711 & 63.63\% & 2.57\% & 85.73\% & 73.05\% & 0.688 \\ \midrule
        \rowcol
        GC1 & - & - & - & 0 & 10 & 57 & 33 & 0.0\% & 14.93\% & 0.0\% & 0.0\% & -0.234 \\ 
        GC2 & - & - & - & 19 & 17 & 50 & 14 & 57.58\% & 25.37\% & 52.78\% & 55.07\% & 0.315 \\ \rowcol
        GC3 & - & - & - & 22 & 34 & 33 & 11 & 66.67\% & 50.75\% & 39.29\% & 49.44\% & 0.151 \\ 
        GC4 & - & - & - & 17 & 31 & 36 & 16 & 51.52\% & 46.27\% & 35.42\% & 41.98\% & 0.049 \\ \rowcol
        GC5 & - & - & - & 20 & 17 & 50 & 13 & 60.61\% & 25.37\% & 54.05\% & 57.14\% & 0.343 \\
        GC6 & - & - & - & 12 & 12 & 55 & 21 & 36.36\% & 17.91\% & 50.0\% & 42.11\% & 0.203 \\ 
        \rowcol
        GC7 & - & - & - & 13 & 14 & 53 & 20 & 39.39\% & 20.9\% & 48.15\% & 43.33\% & 0.196 \\ 
        GC8 & - & - & - & 13 & 11 & 56 & 20 & 39.39\% & 16.42\% & 54.17\% & 45.61\% & 0.253 \\
        \rowcol
        GC9 & - & - & - & 17 & 12 & 55 & 16 & 51.52\% & 17.91\% & 58.62\% & 54.84\% & 0.348 \\
        \midrule
        HP1 & - & - & - & 72 & 66 & 8 & 0 & 100.0\% & 89.19\% & 52.17\% & 68.57\% & 0.237 \\
        \rowcol
        HP2 & - & - & - & 60 & 9 & 65 & 12 & 83.33\% & 12.16\% & 86.96\% & 85.11\% & 0.713 \\ 
        HP3 & - & - & - & 58 & 8 & 66 & 14 & 80.56\% & 10.81\% & 87.88\% & 84.06\% & 0.701 \\ 
        \rowcol
        HP4 & - & - & - & 62 & 10 & 64 & 10 & 86.11\% & 13.51\% & 86.11\% & 86.11\% & 0.726 \\ 
        \midrule
        PD1 & - & - & - & 26 & 8 & 37 & 6 & 81.25\% & 17.78\% & 76.47\% & 78.79\% & 0.63 \\ 
        \rowcol
        PD2 & - & - & - & 11 & 5 & 40 & 21 & 34.38\% & 11.11\% & 68.75\% & 45.83\% & 0.283 \\
        PD3 & - & - & - & 29 & 28 & 17 & 3 & 90.62\% & 62.22\% & 50.88\% & 65.17\% & 0.319 \\ 
        \rowcol
        PD4 & - & - & - & 24 & 6 & 39 & 8 & 75.0\% & 13.33\% & 80.0\% & 77.42\% & 0.623 \\ 
        \bottomrule
    \end{tabular}
\end{table*}
\midsepdefault

\end{document}